\documentclass[10pt]{article} %
\usepackage[accepted]{rlc}

\usepackage{amssymb}            %
\usepackage{mathtools}          %
\usepackage{mathrsfs}           %
\usepackage{graphicx}           %
\usepackage{subcaption}         %
\usepackage[space]{grffile}     %
\usepackage{url}                %
\usepackage{amsmath}
\usepackage{wrapfig}
\usepackage{algorithm}
\usepackage{algpseudocode}
\usepackage{tikz}
\usepackage{graphbox}
\usepackage{arydshln}

\newcommand{\norm}[1]{\left\lVert#1\right\rVert}
\title{Multi-view Disentanglement for Reinforcement Learning with Multiple Cameras}

\author{Mhairi Dunion  \\
    mhairi.dunion@ed.ac.uk \\
    University of Edinburgh
    \And
    Stefano V. Albrecht \\
    s.albrecht@ed.ac.uk\\
    University of Edinburgh}

\begin{document}

\maketitle

\begin{abstract}
The performance of image-based Reinforcement Learning (RL) agents can vary depending on the position of the camera used to capture the images. Training on multiple cameras simultaneously, including a first-person egocentric camera, can leverage information from different camera perspectives to improve the performance of RL. However, hardware constraints may limit the availability of multiple cameras in real-world deployment. Additionally, cameras may become damaged in the real-world preventing access to all cameras that were used during training. To overcome these hardware constraints, we propose Multi-View Disentanglement (MVD), which uses multiple cameras to learn a policy that is robust to a reduction in the number of cameras to generalise to any single camera from the training set. Our approach is a self-supervised auxiliary task for RL that learns a disentangled representation from multiple cameras, with a shared representation that is aligned across all cameras to allow generalisation to a single camera, and a private representation that is camera-specific. We show experimentally that an RL agent trained on a single third-person camera is unable to learn an optimal policy in many control tasks; but, our approach, benefiting from multiple cameras during training, is able to solve the task using only the same single third-person camera.
\end{abstract}

\section{Introduction}
\label{sec:intro}
\begin{wrapfigure}{r}{0.43\textwidth}
	\begin{subfigure}[b]{0.21\textwidth}
            \centering
		\includegraphics[scale=0.17]{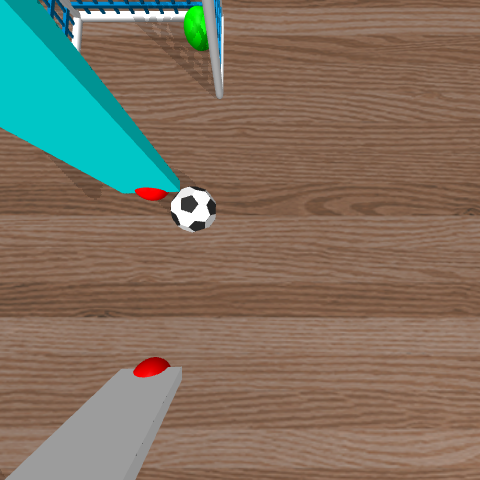}
		\caption{first-person}
            \label{fig:metaworld_first_person}
	\end{subfigure}
        \hfill
	\begin{subfigure}[b]{0.21\textwidth}
		\centering
		\includegraphics[scale=0.17]{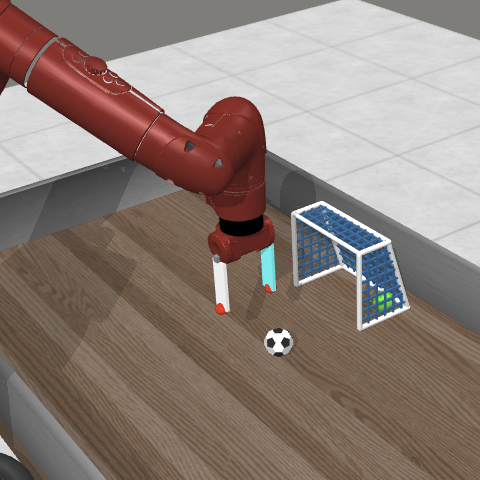}
		\caption{third-person}
            \label{fig:metaworld_third_person}
	\end{subfigure}
	\caption{First-person and third-person camera views for MetaWorld Soccer task.}
\label{fig:metaworld_cams}
\end{wrapfigure}

The ability of a Reinforcement Learning (RL) agent to learn an optimal policy on robotic control tasks from images depends on the position of the camera available during training. Often, a static third-person camera pointing towards the scene (e.g. Figure~\ref{fig:metaworld_third_person}) is not sufficient to learn an optimal policy. A first-person egocentric camera on the robot's end-effector (e.g. Figure~\ref{fig:metaworld_first_person}) has been shown to be necessary to learn an optimal policy in many tasks \citep{Hsu2022VisionManipulators}. Approaches that leverage both first-person and third-person cameras simultaneously have been shown to improve the performance of RL algorithms \citep{Hsu2022VisionManipulators, Jangir2022Look, Barati2019ACAttention}. Whilst it is possible to create multiple camera views in simulation to improve training, access to multiple cameras in the real-world may be restricted due to hardware limitations. Therefore, it is desirable for an RL agent to be able to generalise from multiple cameras to a single camera for successful deployment. Even when multiple cameras are available in the real-world, cameras may fail or become faulty during deployment, so a robust RL agent should continue to perform optimally with only a subset of the cameras that were available during training. We propose to address these hardware limitations with an approach that leverages multiple cameras during training to learn a policy that successfully performs control tasks with multiple cameras as well as achieving robustness to a reduction in the available cameras.

We use multiple cameras to learn a disentangled representation that allows robustness to a reduction in cameras to generalise to any single camera from the training set. Our approach, called \textit{Multi-View Disentanglement} (MVD), is a self-supervised auxiliary task to learn such a disentangled representation and can be used with existing RL algorithms. The learned representation is disentangled into a shared representation that is similar across all cameras, and a private representation that contains information only available to an individual camera. The shared representation gives the RL policy a consistent representation that can be relied upon regardless of the camera. The private representation allows information only available to a single camera to be used during training to improve policy learning as some camera views make it easier to discern important features than others.

We evaluate our approach on robotic control tasks using a Panda robot \citep{Gallouedec2021pandagym, Hsu2022VisionManipulators}, and several MetaWorld tasks \citep{Yu2020metaworld} using a Sawyer robot. We show experimentally that an RL agent often cannot learn an optimal policy when training on a third-person camera alone, and that an approach combining multiple cameras during training \citep{Hsu2022VisionManipulators} does not generalise to a single camera. Our results also show that our approach, MVD, is able to learn an optimal policy from multiple cameras and achieve zero-shot generalisation to successfully solve the task using any of the cameras individually in many tasks.

\section{Related work}
\label{sec:related}

\paragraph{Robotic control with multiple cameras.}
Prior work uses multiple cameras to improve performance on robotic control tasks. \cite{Hsu2022VisionManipulators} use both first-person and third-person cameras together with a variational information bottleneck to regularise the third-person representation, \cite{Jangir2022Look} use transformers with cross-view attention, \cite{Barati2019ACAttention} train multiple workers with different views and combine features for each worker weighted by Q-values, and \cite{Driess2022} use Neural Radiance Fields to learn a representation from multiple images to improve the performance of RL. These approaches combine camera representations to learn a policy that is dependent on all available cameras, and so cannot generalise if one of the training cameras is no longer available. \cite{Acar2023VisualPolicy} trains a teacher RL policy on multiple cameras augmented with human demonstrations. They use imitation learning to learn a student single-camera policy to output an action similar to the teacher multi-camera policy. \cite{Shang2021SelfSupervisedDR} also use imitation learning where demonstrations are a third-person view from a human or robot, which are used alongside first-person views of the egocentric robot to learn a disentangled representation for imitation learning. Our approach does not require training a separate teacher policy or collecting suitable demonstrations, and instead learns a disentangled representation online that allows generalisation to a single camera.

\paragraph{Multi-view representation learning.}
Multi-view representation learning approaches use multiple sources/views of a shared context. This can consist of multiple camera views of the same scene (as is the case in our work) as well as combining multi-modal inputs. 
\cite{Li2019survey} categorise approaches into representation alignment and representation fusion. Representation alignment includes minimising the distance between representations of different views \citep{Feng2014, Li2003}, maximising similarity between views \citep{Bachman2019MaxMI, Frome2013DeViSE} and maximising correlation of variables across views \citep{Andrew2013CCA}. Representation fusion combines the representations from different views into a single representation for downstream tasks \citep{Geng2022MultimodalMA, Xie2021, Karpathy2014DeepVA}. 
Multi-view disentanglement separates the learned representation into shared and private parts. The shared representation is aligned across views, while the private representation is view-specific. Several approaches have been proposed to achieve multi-view disentanglement in the supervised and unsupervised learning literature \citep{Jain2023, Ke2023, Xu2021multiVAE, Ke2021conan, Xin2021, Hosoya2019, GonzalezGarcia2018ImagetoimageTF, Ye2016LearningMV}.
Our work is aligned with the multi-view disentanglement literature, but we consider multi-view disentanglement for RL and use the temporal data available in RL.

\paragraph{Self-supervised auxiliary tasks in RL.}
Disentangled representations have been used to improve RL generalisation for a single camera \citep{Dunion2023cmid, Dunion2023ted, Higgins2017darla}. Other approaches learn representations using mutual information \citep{Lee2020pisac, garcin2024dred}, regression targets \citep{mcinroe2023hksl} and similarity constraints \citep{Agarwal2021, Mazoure2020, Oord2018Contrastive}. CURL \citep{Laskin2020Curl} uses contrastive learning to maximise the similarity between representations of the same image with different augmentations to improve sample efficiency. However, \cite{li2022does} find that self-supervised learning frameworks with augmented images have limited impact on RL performance compared to image augmentation alone. While we also use contrastive learning with images, we consider camera views rather than augmentations.

\section{Preliminaries}
\label{sec: prelim}

\paragraph{Reinforcement learning.}
We assume the agent is acting in a Markov Decision Process (MDP), defined by the tuple ${\mathcal{M} = (\mathcal{S}, \mathcal{A}, P, R, \gamma)}$, where $\mathcal{S}$ is the state space, $\mathcal{A}$ is the action space, 
$P(x_{t+1}|x_t,a_t)$ is the probability of next state $x_{t+1} \in \mathcal{S}$ given action $a_t \in \mathcal{A}$ is taken in state $x_t \in \mathcal{S}$ at time $t$,
$R(x_t, a_t)$ is the reward function giving reward $r_t$ after taking action $a_t$ in state $x_t$, and $\gamma \in [0,1)$ is the discount factor. 
The goal of an RL agent is to learn a policy $\pi$ to maximise the discounted return, $\max_{\pi} \mathbb{E}_{P, \pi} [\sum_{t=0}^{\infty}[\gamma^t R(x_t, a_t)]]$. 
In this work, we focus on RL from image pixels, where the agent observation $\mathbf{o}_t \in \mathbb{R}^{c \times h \times w}$ at timestep $t$ is an RGB image, a high-dimensional observation of the underlying state, where $c$ is the channels, and $h$ and $w$ are the image height and width respectively. An observation can consist of multiple consecutive image frames where required for the task. The agent learns a lower-dimensional latent representation $\mathbf{z}_t$ and the policy $\pi$ is now a function of the learned representation, $\pi(\mathbf{a}_t | \mathbf{z}_t)$. We consider $N$ different camera views of the same scene; we use $C$ to denote the set of cameras and $c_i \in C$ is a single camera. We use superscript to identify the observation $\mathbf{o}_t^{c_i}$ or representation $\mathbf{z}^{c_i}_t$ for camera $c_i \in C$.

\paragraph{Contrastive learning.}
We use contrastive learning to learn disentangled representations in a self-supervised way. Given a query $\textbf{q}$ (sometimes also referred to as an anchor), contrastive learning aims to maximise the similarity between $\textbf{q}$ and a positive key $\textbf{k}^+$ while minimising the similarity with each negative key $\textbf{k}^-$. Many approaches use the dot product $\textbf{q}^T\textbf{k}$ to measure the similarity between vectors $\textbf{q}$ and $\textbf{k}$ \citep{Chen2020SimCLR, He2020MoCo, Wu2018} or the bilinear product $\textbf{q}^TW\textbf{k}$ where $W$ is a learnable weight matrix \citep{Laskin2020Curl, Oord2018Contrastive}.

We use the InfoNCE loss \citep{Oord2018Contrastive} with normalised dot product similarity measure, also known as cosine similarity, to measure the distance between vectors. This loss function was also used for the SimCLR \citep{Chen2020SimCLR} and MoCo \citep{He2020MoCo} methods. Let $\text{sim}(\textbf{q},\textbf{k})$ denote the cosine similarity between two vectors $\textbf{q}$ and $\textbf{k}$, given by:
\begin{equation}
\label{eq:sim}
    \text{sim}(\textbf{q},\textbf{k}) = \frac{\textbf{q}^T\textbf{k}}{\norm{\textbf{q}}\norm{\textbf{k}}}.
\end{equation}
Then the InfoNCE loss with query $\textbf{q}$, positive key $\textbf{k}^+$ and $M$ negative keys $\{ \textbf{k}_i^- \}_{i=1}^M$ is given by:
\begin{equation}
\label{eq:infonce}
    \mathcal{L}^{\text{InfoNCE}}(\textbf{q}, \textbf{k}^+, \{ \textbf{k}_i^- \}_{i=1}^M) = \log \left[ \frac{\exp{(\text{sim}(\textbf{q}^T,\textbf{k}^+)/\tau)}}{\exp{(\text{sim}(\textbf{q}^T,\textbf{k}^+)/\tau)} + \sum_{i=1}^{M}\exp{(\text{sim}(\textbf{q}^T,\textbf{k}_i^-)/\tau)}}  \right]
\end{equation}
where $\tau$ is a temperature parameter. 

\section{Multi-view disentanglement for RL}
\label{sec:approach}
We propose \textit{Multi-View Disentanglement} (MVD) to learn a disentangled representation from multiple cameras that is robust to a reduction in available cameras. Our approach learns separate shared $\mathbf{s}_t^{c_i}$ and private $\mathbf{p}_t^{c_i}$ representations for each camera $c_i \in C$, where $C$ denotes the set of cameras and $t$ is the timestep. The shared representation $\mathbf{s}_t^{c_i}$ is trained to be similar across all cameras $\forall c_i \in C$, while the private representation $\mathbf{p}_t^{c_i}$ encodes information available only to a specific camera $c_i \in C$. Both representations together are used to condition the RL policy $\pi(\mathbf{a}_t | \mathbf{z}_t)$ with $\mathbf{z}_t = (\mathbf{s}_t^{c_i}, \mathbf{p}_t^{c_j} )$, where the shared part allows for camera generalisation and the private part allows the policy to leverage extra information available to a single camera to improve training. We define two contrastive learning objectives that together give an auxiliary task for multiple camera views that can be applied to existing RL algorithms. We first provide an overview of the approach in Section~\ref{subsec:overview}, then discuss the MVD auxiliary task in Section~\ref{subsec:disentanglement}. Finally, we explain how the disentangled representation is used as input to the RL policy in Section~\ref{subsec:rl}.

\begin{figure}
    \centering
    \includegraphics[scale=0.8]{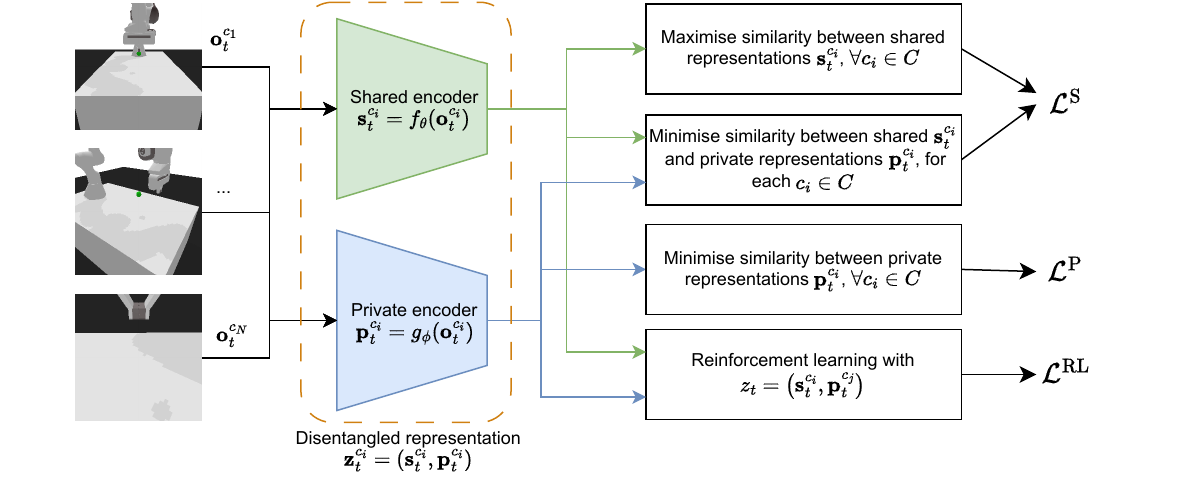}
    \caption{Multi-view Disentanglement (MVD) architecture. Each camera image is used to generate a shared and private representation. The shared auxiliary loss $\mathcal{L}^{\text{S}}$ uses these representations to maximise similarity between shared representations and minimise similarity between shared and private representations. The private auxiliary loss $\mathcal{L}^{\text{P}}$ minimises similarity between private representations.}
    \label{fig:architecture}
\end{figure}

\subsection{Overview of approach}
\label{subsec:overview}
An outline of our approach is depicted in Figure~\ref{fig:architecture}. MVD consists of two separate encoders, which both learn a lower-dimensional latent representation of the same size for each camera $c_i \in C$. There is one encoder for the shared representation, $\mathbf{s}^{c_i}_t = f_{\theta}(\mathbf{o}_t^{c_i})$, where $f_{\theta}$ is an encoder parameterised by $\theta$. There is a separate encoder for the private representation, $\mathbf{p}^{c_i}_t = g_{\phi}(\mathbf{o}_t^{c_i})$, where $g_{\phi}$ is an encoder parameterised by $\phi$. Both encoders take the same observation image pixels of a single camera, $\mathbf{o}_t^{c_i}$, as input and encoder parameters are shared across all cameras. We use a contrastive learning approach to maximise similarity between the shared representations, $\mathbf{s}^{c_i}_t$, of all cameras $c_i \in C$ while minimising similarity between the shared $\mathbf{s}^{c_i}_t$ and private $\mathbf{p}^{c_i}_t$ representations for each camera $c_i \in C$ to achieve disentanglement. An additional contrastive loss is used to minimise similarity between private representations $\mathbf{p}^{c_i}_t$ for all $c_i \in C$ to ensure the policy cannot rely solely on the private representation. The contrastive learning details are provided in Section~\ref{subsec:disentanglement}. Both the shared and private representations are used as input to the RL algorithm, which will be described in Section~\ref{subsec:rl}.

\subsection{Multi-view disentanglement}
\label{subsec:disentanglement}
\paragraph{Shared contrastive loss.}
The goal of the shared contrastive loss is to disentangle the shared $\textbf{s}^{c_i}_t$ and private $\textbf{p}^{c_i}_t$ representations for each camera $c_i \in C$ by minimising the similarity between these representations, while also maximising the similarity between the shared representation $\textbf{s}_t^{c_i}$ for all cameras $\forall c_i \in C$ to achieve alignment. These two objectives are combined into one contrastive loss by defining suitable positive and negative keys. We use the InfoNCE loss in Equation~\ref{eq:infonce} with the cosine similarity measure in Equation~\ref{eq:sim}. For each calculation of the InfoNCE loss, one camera $c_q \in C$ is selected as the query camera and another as the positive camera $c_+ \in C$ where $c_q \neq c_+$. The shared representation for the query camera $c_q$ is the query for the contrastive loss, $\textbf{q} = \textbf{s}_t^{c_q}$. The shared representation of the positive camera $c_+$ is the positive key, $\textbf{k}^+ = \textbf{s}_t^{c_+}$. The negative keys consist of two sets: (1) the positive keys of all other queries in the batch $\{\textbf{s}_{t'}^{c_+}\}_{t' \neq t}$, and (2) the private representations for all $N$ cameras $\{\textbf{p}_t^{c_n}\}_{n=1}^{N}$. The positive key encourages the shared representations to be similar across cameras, while the negative keys encourage (1) shared representations to be different for different timesteps to capture task-relevant information, and (2) shared and private representations to be dissimilar to disentangle them. The query, positive and negative keys are used as the input to the InfoNCE loss:

\begin{equation}
    \mathcal{L}^{\text{S}} = \sum_{c_q \neq c_+} \mathcal{L}^{\text{InfoNCE}}(\textbf{s}_t^{c_q}, \textbf{s}_t^{c_+}, \{\textbf{s}_{t'}^{c_+}\}_{t' \neq t} \cup \{\textbf{p}_t^{c_n}\}_{n=1}^{N})
\label{eq:mvd_shared}
\end{equation}
where we sum the loss over each camera in a single update step.

\paragraph{Private contrastive loss.}
To prevent the shared representation from collapsing and the agent relying solely on the private representation, we minimise the similarity between the private representations. We use the InfoNCE loss with cosine similarity again to achieve this. The query is the private representation for a given camera at timestep $t$, $\mathbf{q} = \mathbf{p}_t^{c_q}$ for $c_q \in C$. The positive key is the private representation for the same camera at the next timestep $t+1$, such that $\mathbf{k}^+ = \mathbf{p}_{t+1}^{c_q}$. The negative keys are the private representations for all other cameras at timestep $t$, given by $\{\mathbf{p}_{t}^{c_n}\}_{\forall n\neq q}$. The positive keys are needed to keep the representation bounded and are chosen to encourage temporal consistency in the private representations at consecutive timesteps, while the negative keys encourage the private representation for each camera to be dissimilar. This prevents the agent from encoding information available to all cameras in the private representation. The query, positive and negative keys are used as the input to the InfoNCE loss:

\begin{equation}
    \mathcal{L}^{\text{P}} = \sum_{c_q} \mathcal{L}^{\text{InfoNCE}}(\textbf{p}_t^{c_q}, \textbf{p}_{t+1}^{c_q}, \{\textbf{p}_t^{c_n} \}_{\forall n \neq q})
\end{equation}

\paragraph{MVD loss.}
Since the similarity measure is normalised, we combine the shared and private contrastive losses by summing them, giving the MVD loss:
\begin{equation}
    \mathcal{L}^{\text{MVD}} = \mathcal{L}^{\text{S}} + \mathcal{L}^{\text{P}}
\end{equation}

\subsection{Reinforcement learning}
\label{subsec:rl}
We use both the shared and private representations to create the representation $\mathbf{z}_t$ as input to the RL algorithm (for both value and policy where applicable) because the shared representation alone does not benefit from the features that are easier to discern or only available to one camera during training. One shared representation $\mathbf{s}_t^{c_i}$ and private representation $\mathbf{p}_t^{c_j}$ are randomly sampled for each update step, such that $\mathbf{z}_t = \left(\mathbf{s}_t^{c_i}, \mathbf{p}_t^{c_j} \right)$ with $c_i, c_j \in C$.
The shared and private representation do not have to come from the same camera due to the consistency across shared representations.
We randomly sample the representations at each update step because combining all representations would result in a policy that can learn to rely on all cameras and would therefore fail to generalise if one camera is unavailable. Both the RL and MVD losses are used to update both shared and private encoders. Combining the RL loss with the MVD loss gives the loss used in training at every update step:
\begin{equation}
    \mathcal{L} = \mathcal{L}^{\text{RL}} + \mathcal{L}^{\text{MVD}}
\end{equation}
where $\mathcal{L}^{\text{RL}}$ is the RL loss for the chosen RL algorithm.

\section{Experimental results}
Our experiments evaluate performance on multiple cameras as well as zero-shot generalisation to any single camera from the training set. We evaluate our approach on robotic control tasks with a Panda robot using the Reach task from Panda Gym \citep{Gallouedec2021pandagym} and a cube grasping task from \cite{Hsu2022VisionManipulators}. We also evaluate on four MetaWorld \citep{Yu2020metaworld} tasks using a Sawyer robot. Our results show that MVD learns a policy that can solve the task with multiple cameras and generalise to a single camera. In many experiments, our approach is able to learn an optimal policy for a single camera where an agent trained directly on that camera alone is unable to learn at all. 

\begin{wrapfigure}{r}{0.47\textwidth}
	\begin{subfigure}[t]{0.15\textwidth}
            \centering
		\includegraphics[scale=0.25]{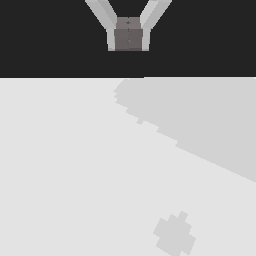}
            \caption{first-person}
	\end{subfigure}
	\hfill
	\begin{subfigure}[t]{0.15\textwidth}
		\centering
		\includegraphics[scale=0.25]{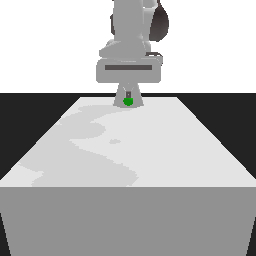}
		\caption{third-person (front)}
	\end{subfigure}
        \hfill
        \begin{subfigure}[t]{0.15\textwidth}
		\centering
		\includegraphics[scale=0.25]{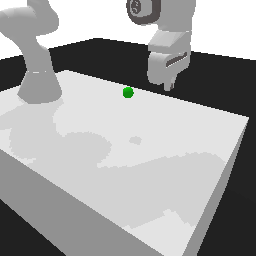}
		\caption{third-person (side)}
	\end{subfigure}
	\caption{Camera views used for Panda tasks.}
\label{fig:panda_cams}
\end{wrapfigure}

\begin{figure}
\centering
\begin{subfigure}{1.0\textwidth}
\begin{tabular}{cc:ccc}
    & all cameras & first-person & third-person (front) & third-person (side) \\
    \rotatebox[origin=c]{90}{Reach} & \includegraphics[align=c, scale=0.17]{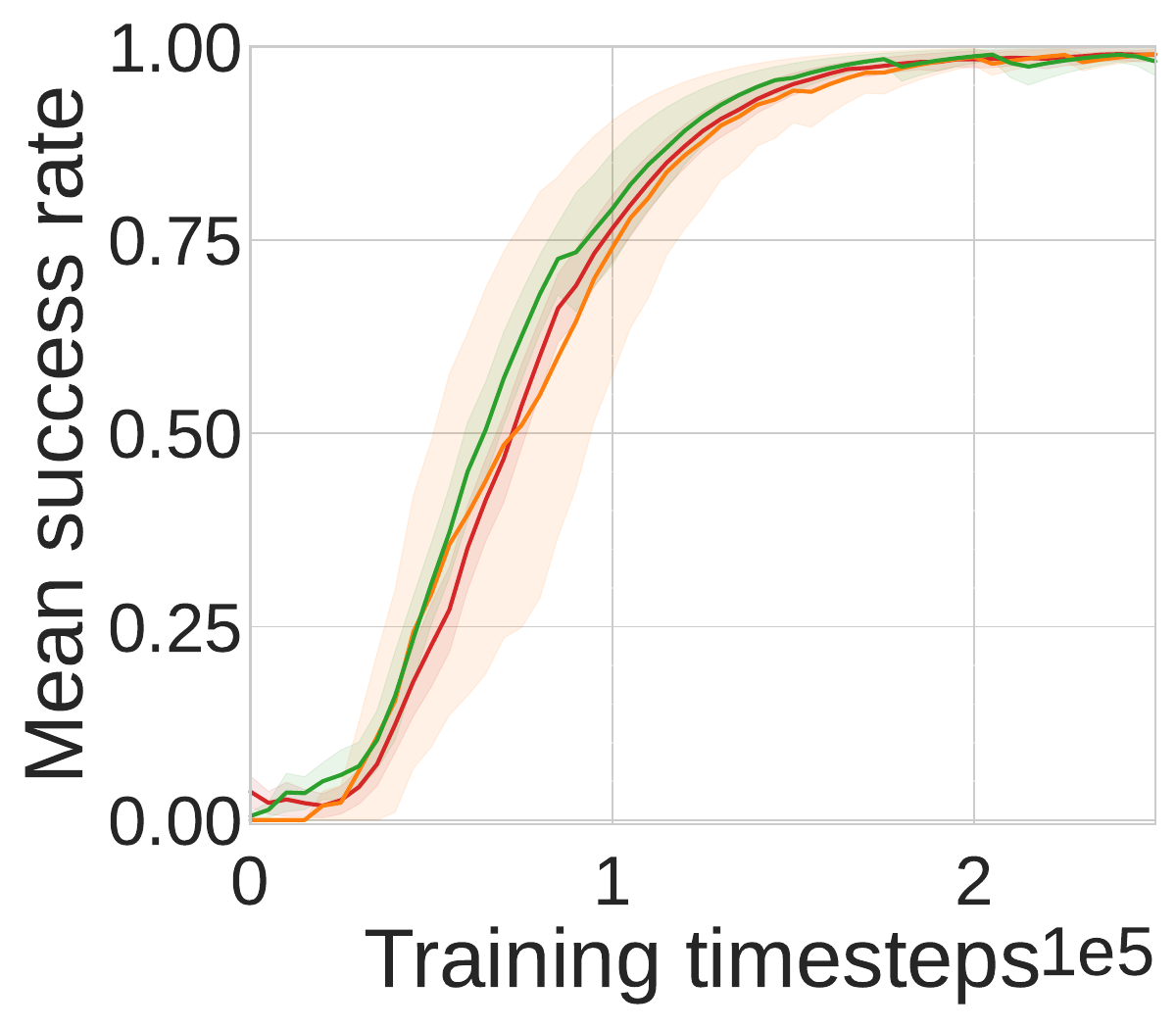} & \includegraphics[align=c, scale=0.16]{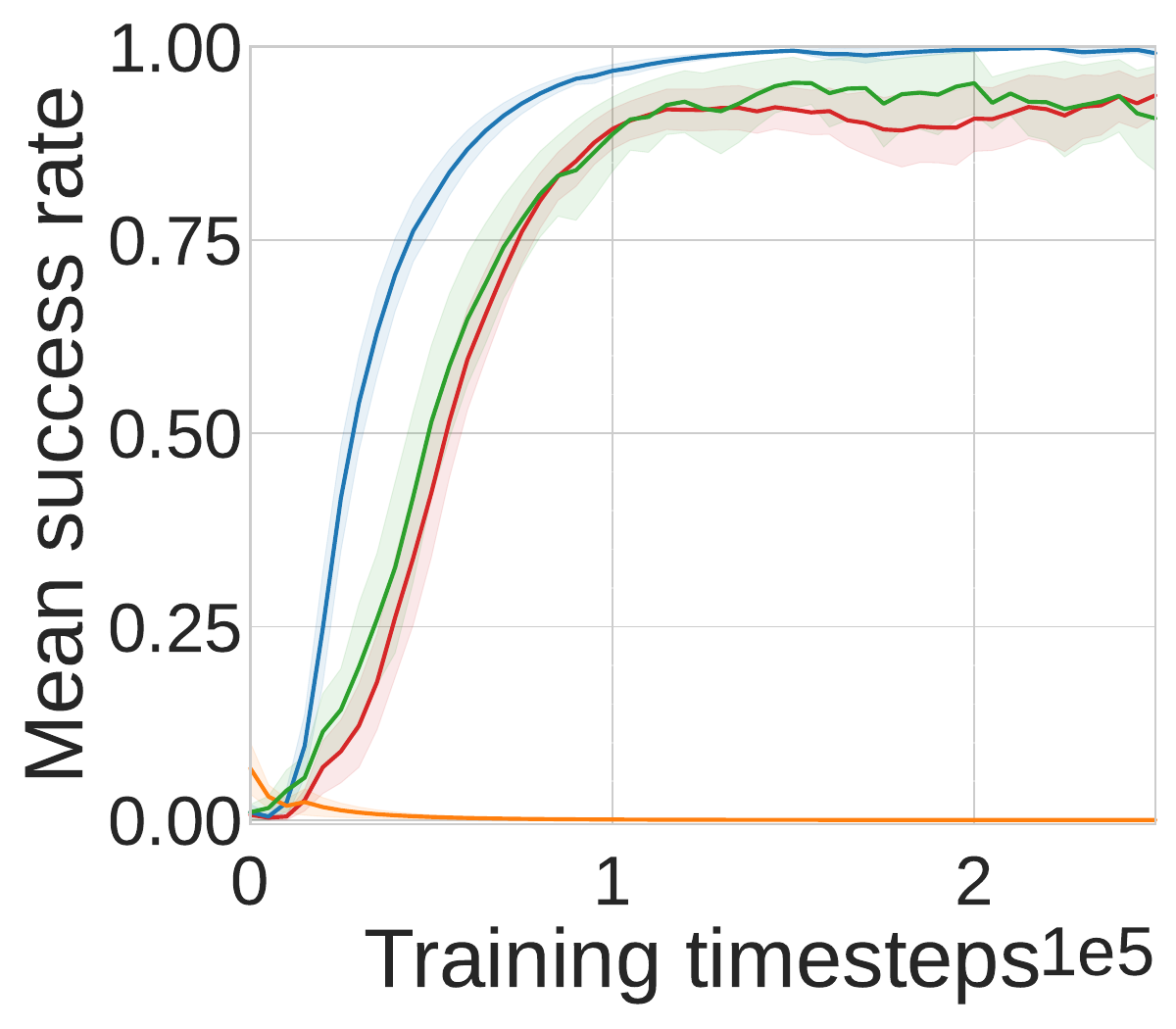} & 		
    \includegraphics[align=c, scale=0.16]{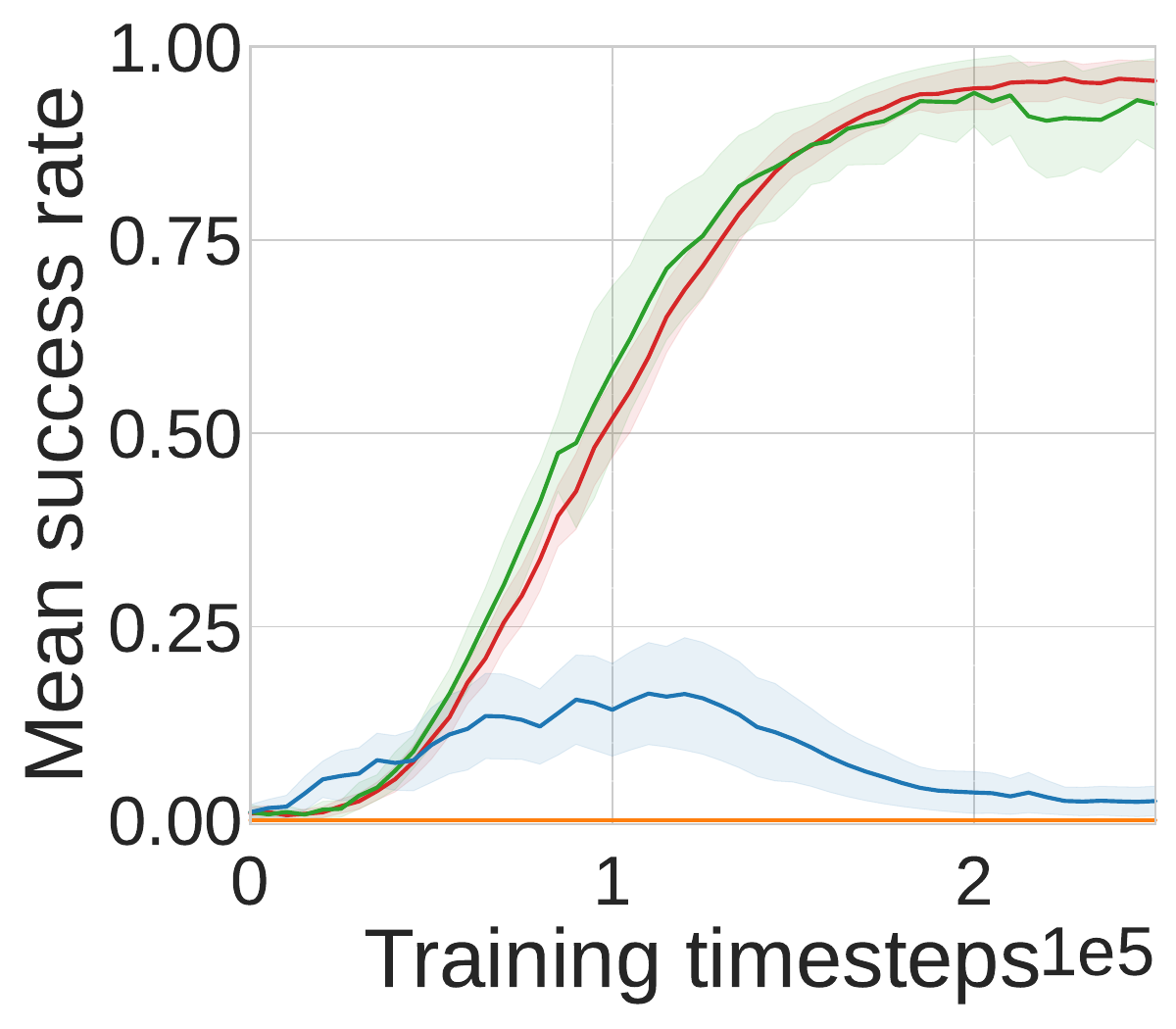} &
    \includegraphics[align=c, scale=0.16]{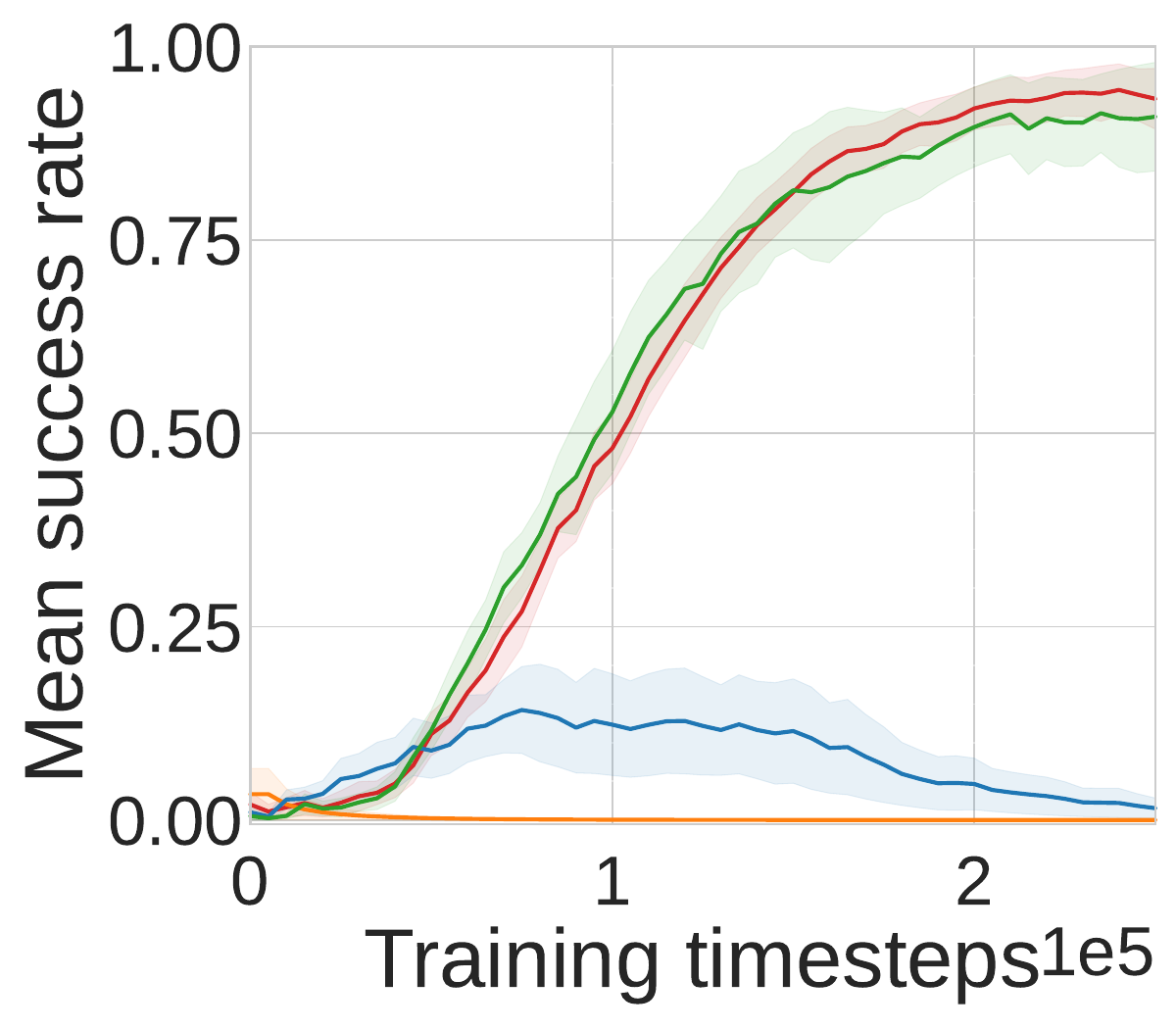} \\
    \rotatebox[origin=c]{90}{Cube Grasping} & \includegraphics[align=c, scale=0.17]{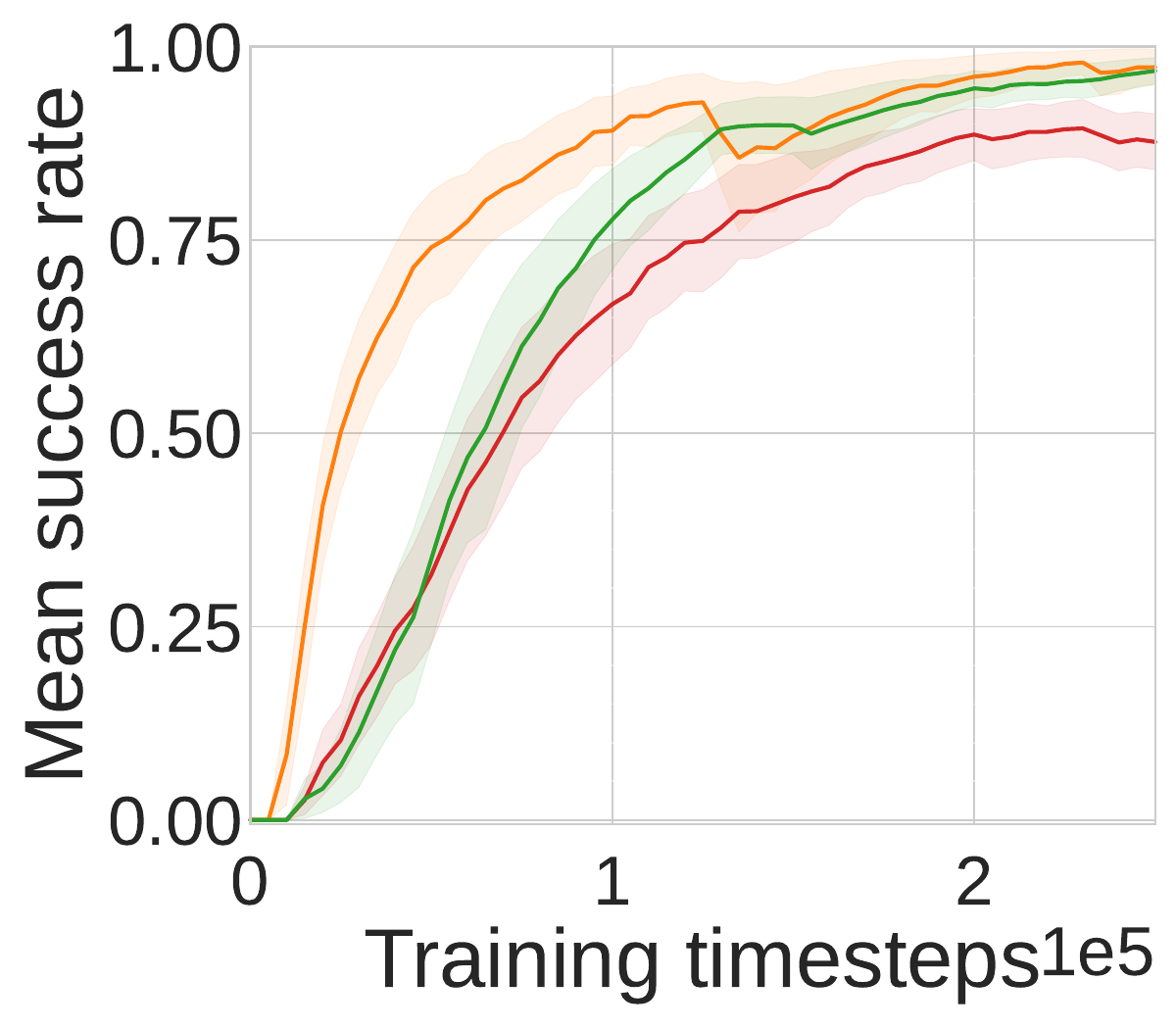} & \includegraphics[align=c, scale=0.16]{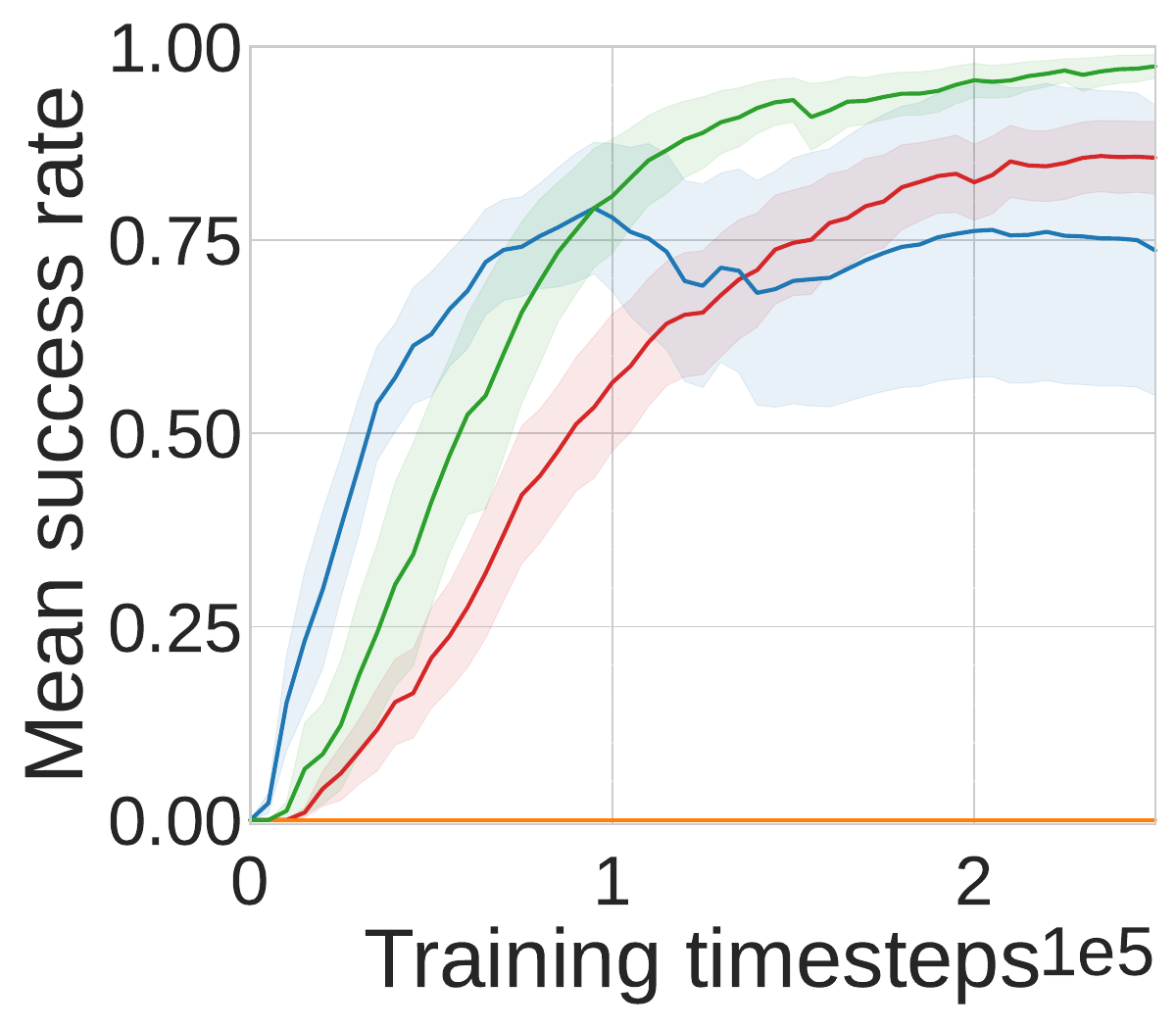} & 		
    \includegraphics[align=c, scale=0.16]{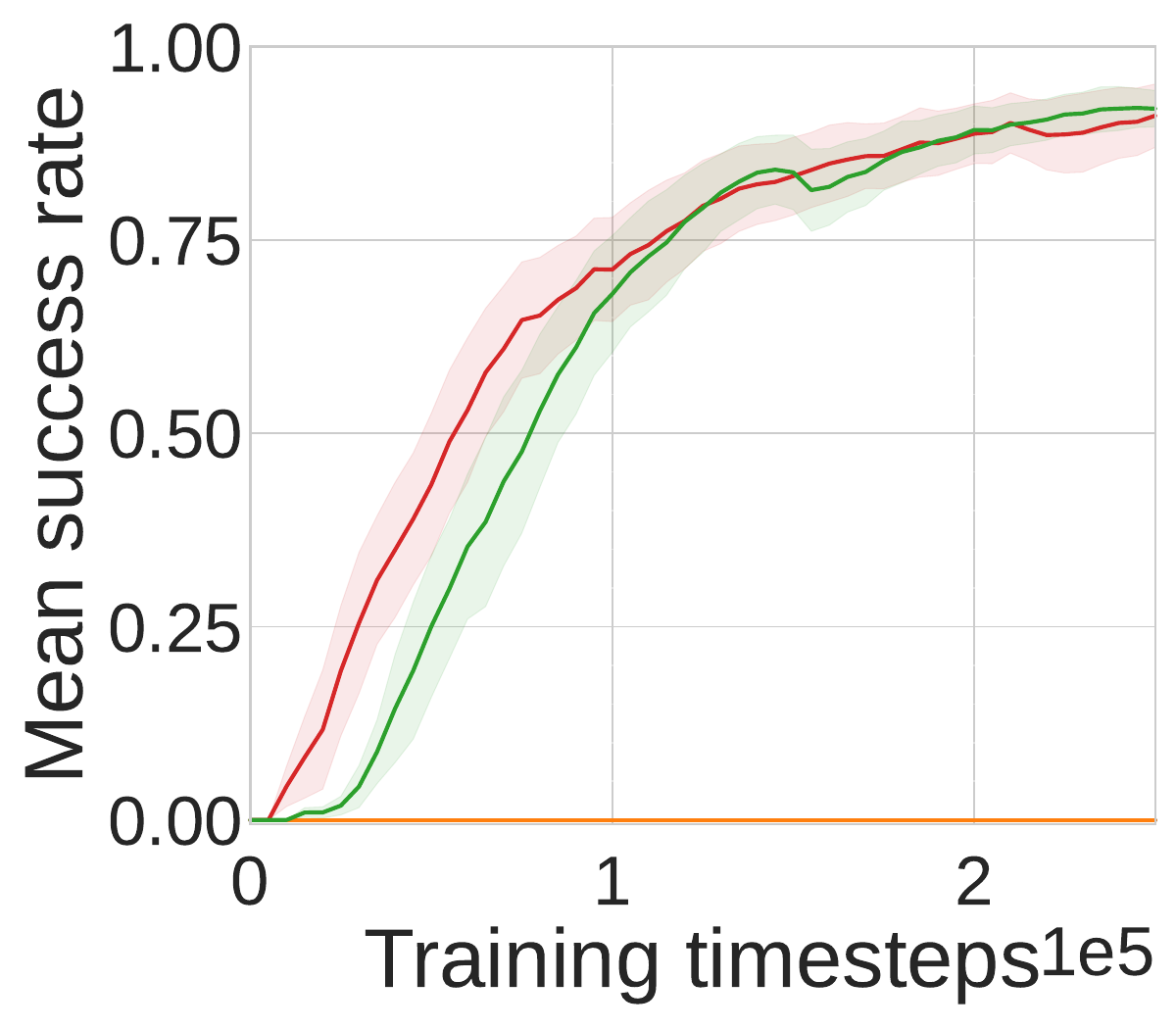} &
    \includegraphics[align=c, scale=0.16]{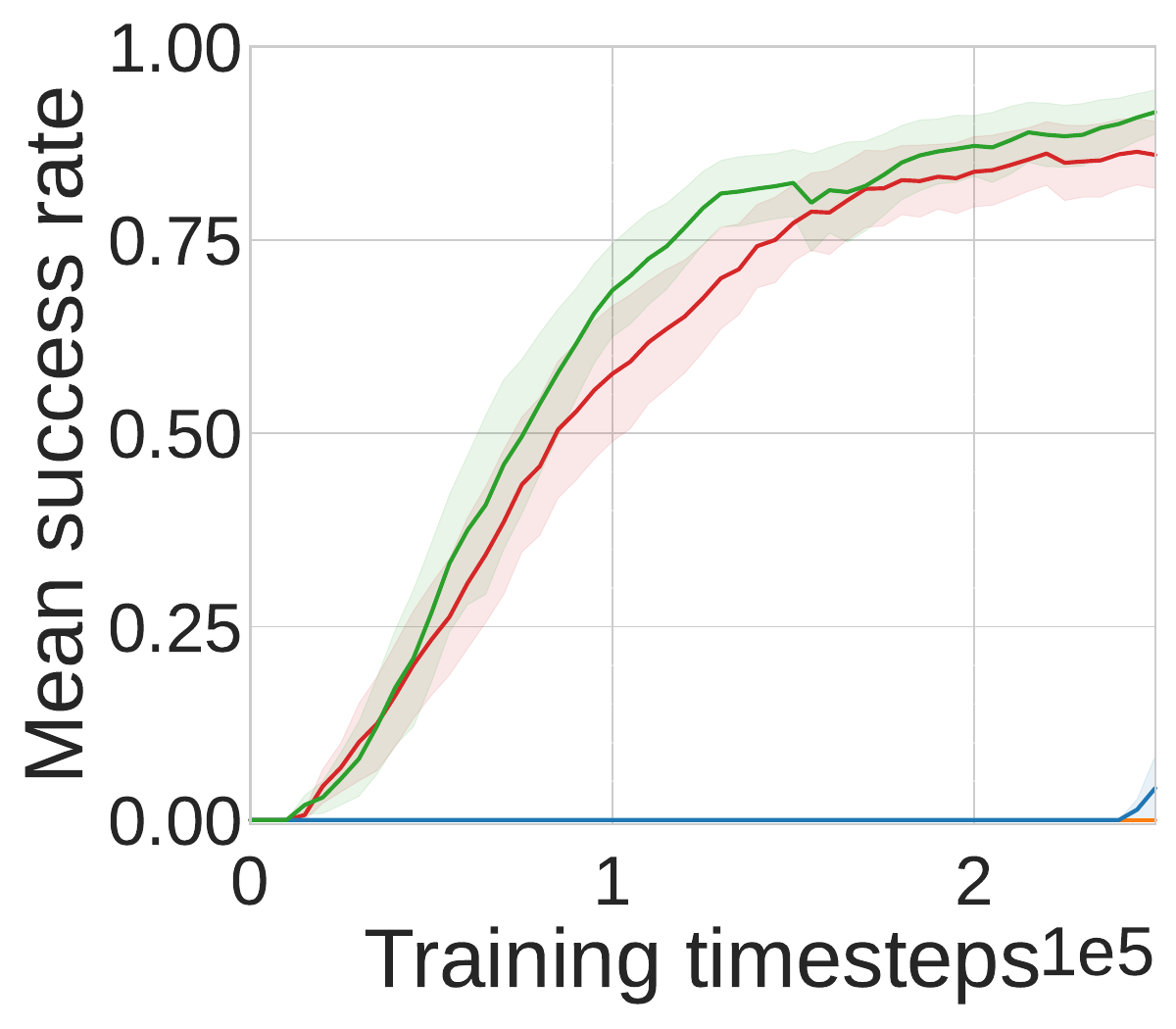}
\end{tabular}
\end{subfigure}
\begin{subfigure}{1.0\textwidth}
\centering
\vspace{2mm}
\includegraphics[scale=0.25]{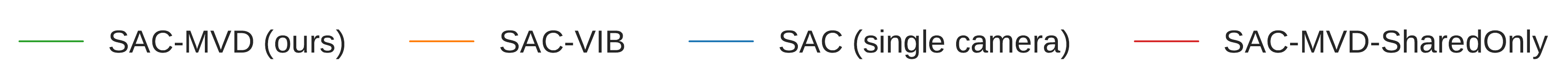}
\end{subfigure}
\caption{Results for Panda tasks showing success rate for evaluation on all cameras (left of dashed line) compared with success rate on each of the individual cameras (right). Success rate is averaged over 20 evaluation episodes for 5 seeds. The shaded region is standard deviation.}
\label{fig:panda_results}
\end{figure}

\subsection{Experimental setup}
To demonstrate the broad applicability of our approach, we apply MVD to two RL algorithms and different numbers of cameras. For the Panda tasks we use SAC \citep{Haarnoja2018sac} with a decoder to aid learning from images and three cameras (depicted in Figure~\ref{fig:panda_cams}). For the MetaWorld tasks we use DrQ \citep{Yarats2021drq}, which uses image augmentations to improve sample efficiency on these more difficult tasks, and two cameras (depicted in Figure~\ref{fig:metaworld_cams}). %
In all tasks, one of the cameras is a first-person camera on the end-effector, while the remaining views are static third-person cameras. Implementation details and hyperparameters are provided in Appendix~\ref{appendix:implementation}. Our code is available at \href{https://github.com/uoe-agents/MVD}{github.com/uoe-agents/MVD}.

In the MetaWorld tasks only, we also concatenate the representation with proprioceptive state information to improve training, similar to prior work using MetaWorld from images \citep{Hsu2022VisionManipulators}. The proprioceptive state consists of the 3D end-effector position and 1D gripper width, and is used for baselines as well as our method. However, we also demonstrate that our approach does not depend on proprioceptive state information by excluding it on the Panda tasks. 

To demonstrate the importance of using multiple cameras in training, we compare to the same base RL algorithm trained only on a single camera for each available camera separately. We also demonstrate the importance of our disentanglement approach for generalisation over representation fusion approaches by comparing to VIB~\citep{Hsu2022VisionManipulators}, an approach that combines camera views with a variational information bottleneck on the third-person camera, for which we use the same base RL algorithm as MVD in each task. Finally, we include an ablation of our method that has only the shared representation with the corresponding loss to maximise similarity between the shared representation for all cameras (`MVD-SharedOnly'). This ablation is used to demonstrate the importance of the private representation during training.

\begin{figure}[t]
\begin{subfigure}{1.0\textwidth}
\centering
\begin{tabular}{cc:cc}
    & all cameras & first-person & third-person \\
    \rotatebox[origin=c]{90}{Soccer} & \includegraphics[align=c, scale=0.17]{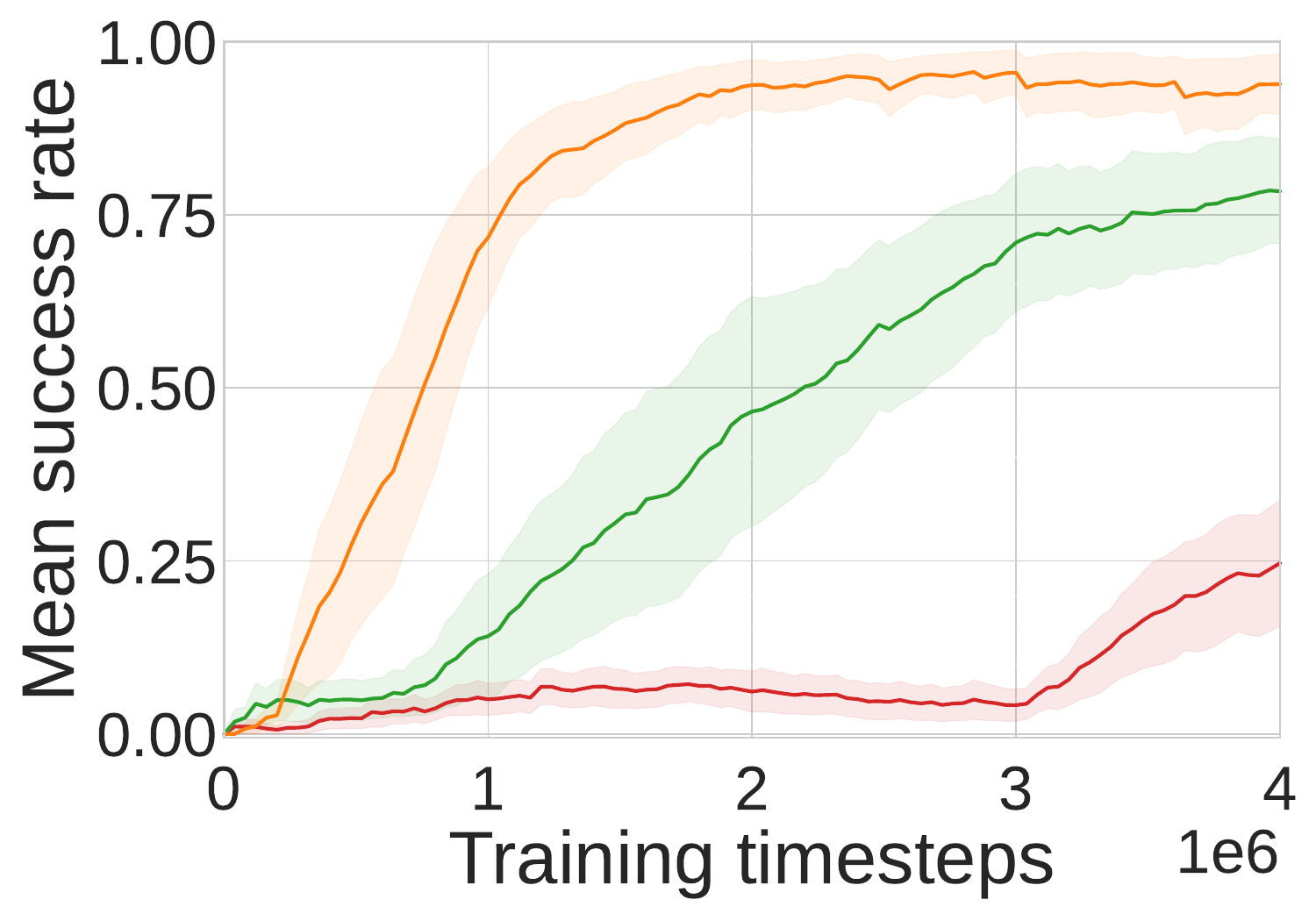} & \includegraphics[align=c, scale=0.17]{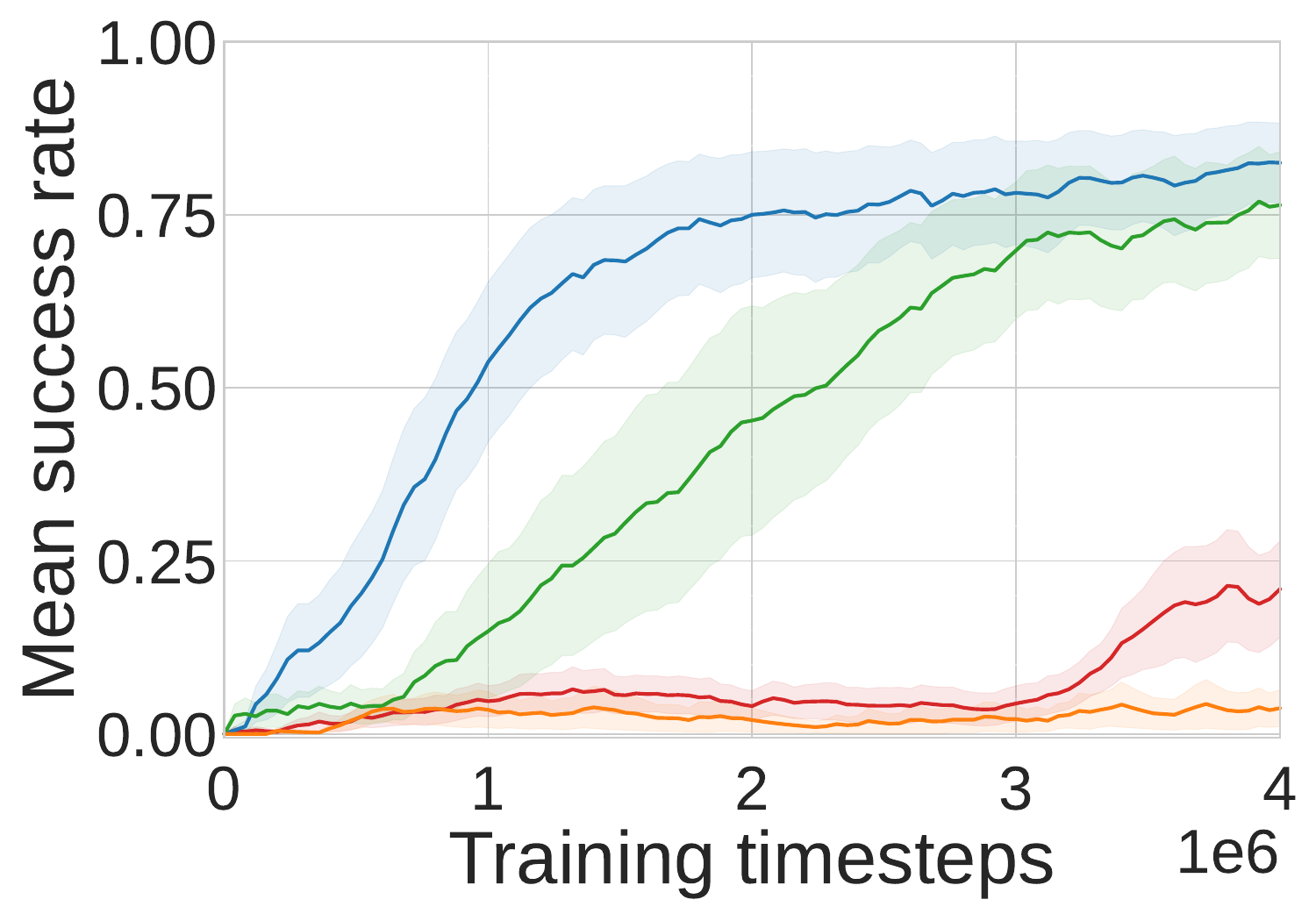} & 		\includegraphics[align=c, scale=0.17]{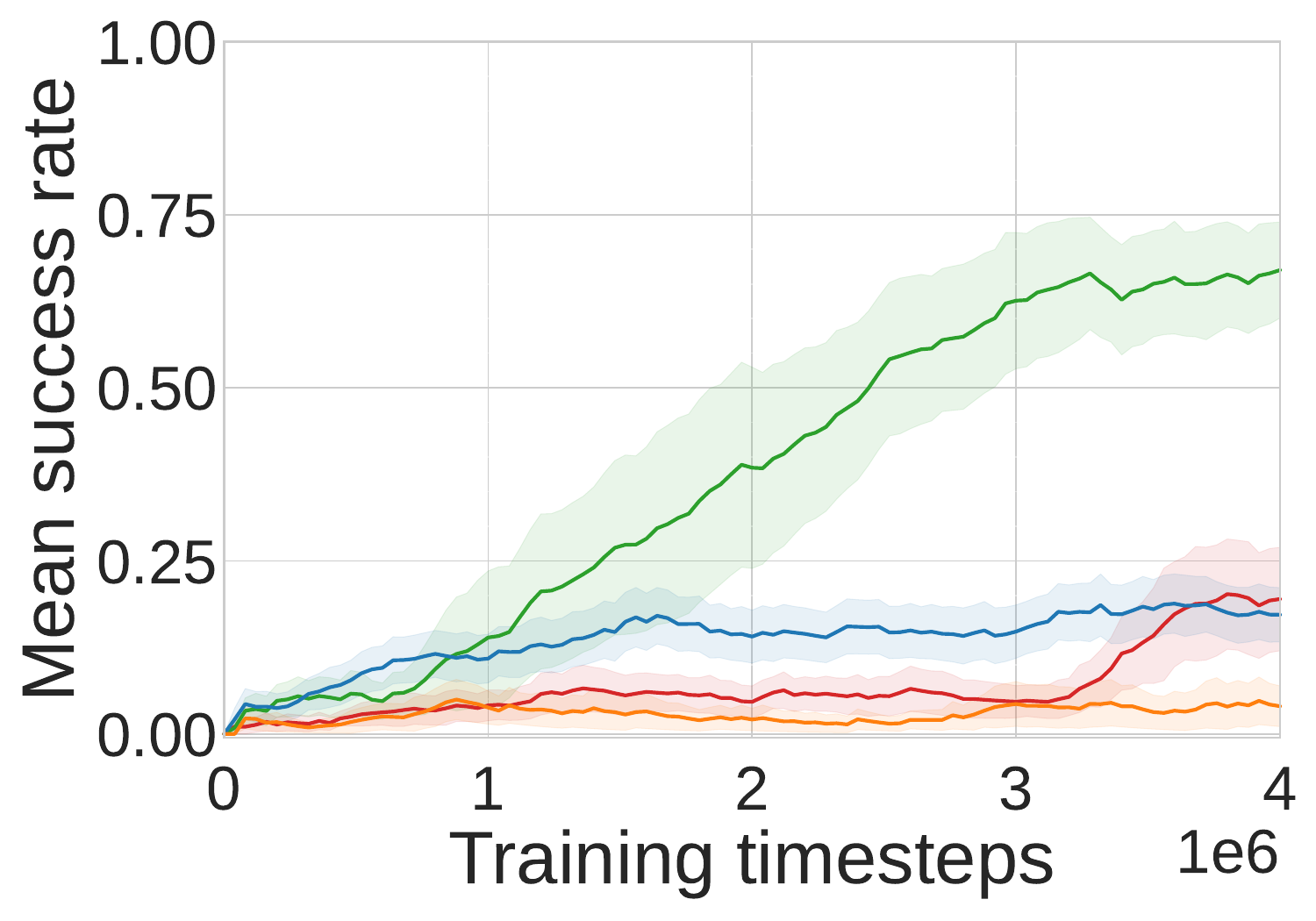} \\
    \rotatebox[origin=c]{90}{Basketball} & \includegraphics[align=c, scale=0.17]{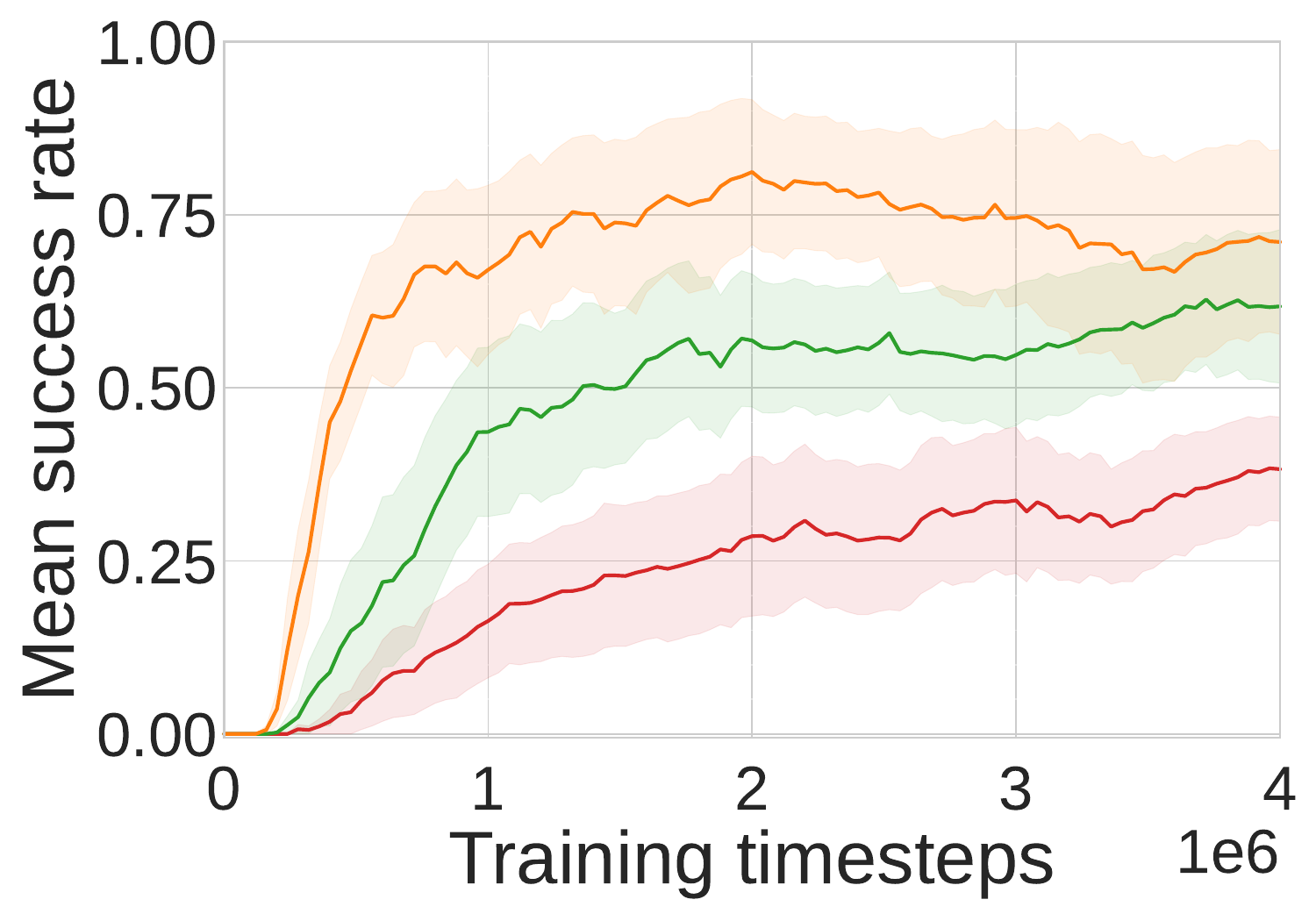} & \includegraphics[align=c, scale=0.17]{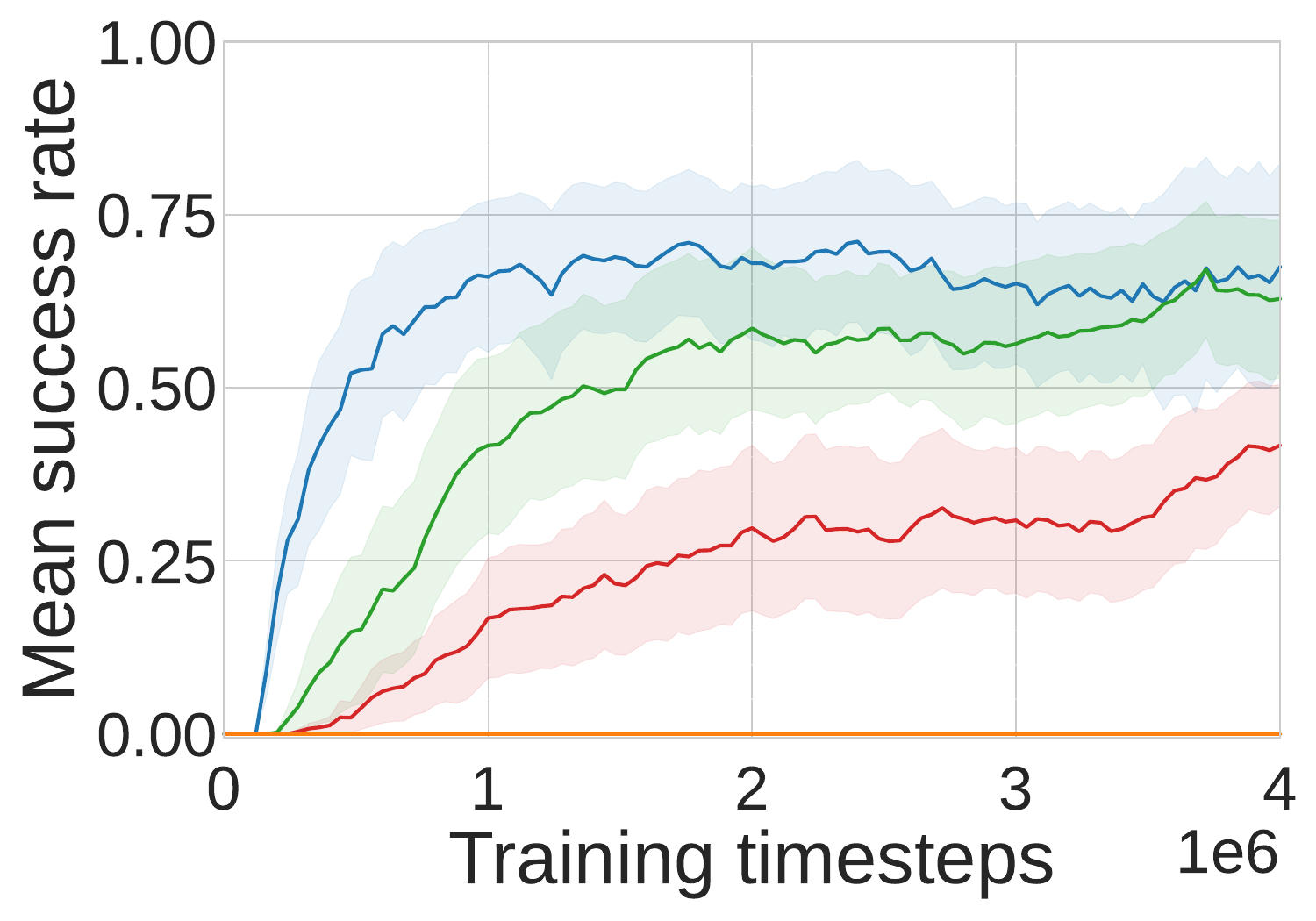} & 		\includegraphics[align=c, scale=0.17]{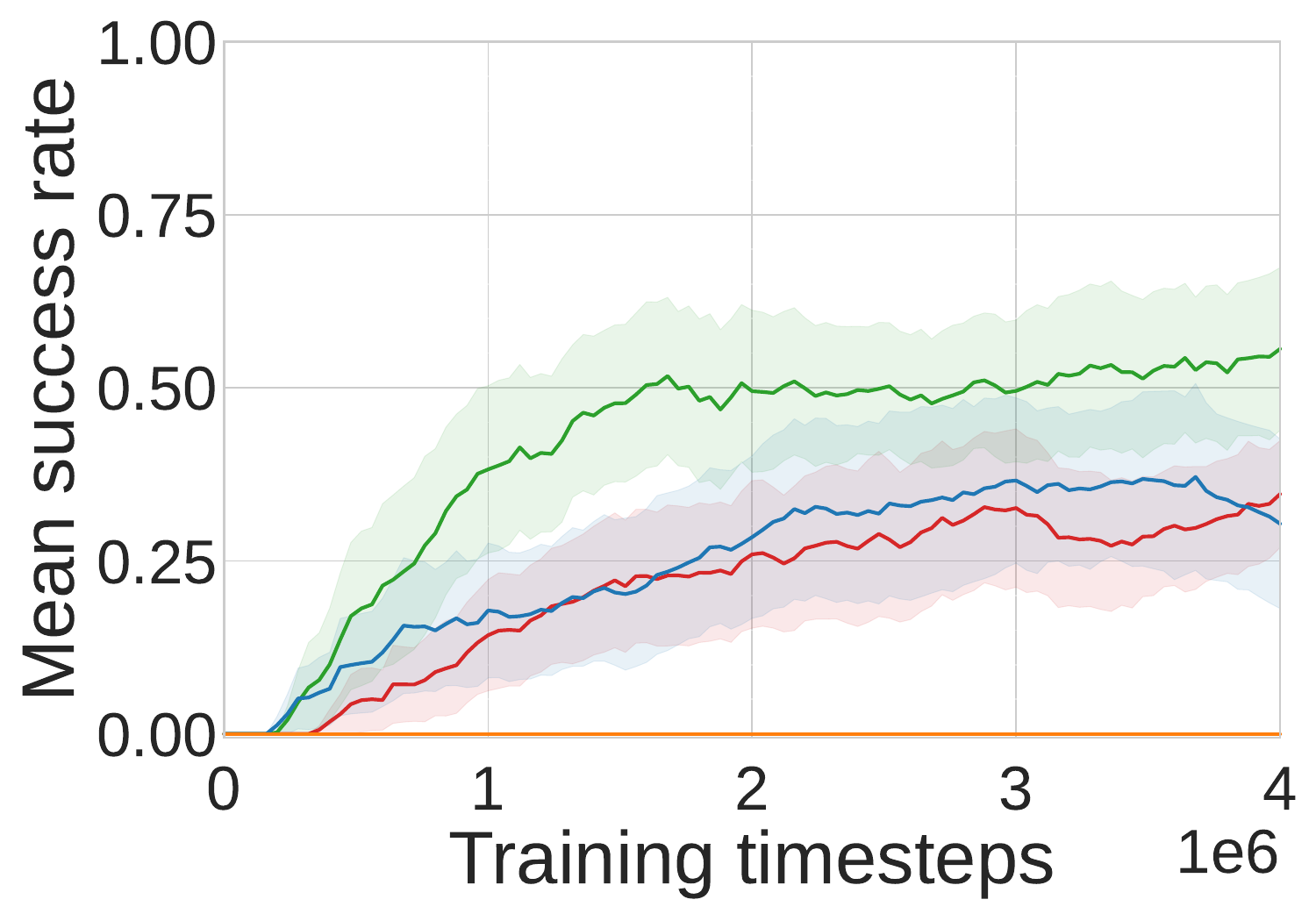} \\
   \rotatebox[origin=c]{90}{Pick and Place} & \includegraphics[align=c, scale=0.17]{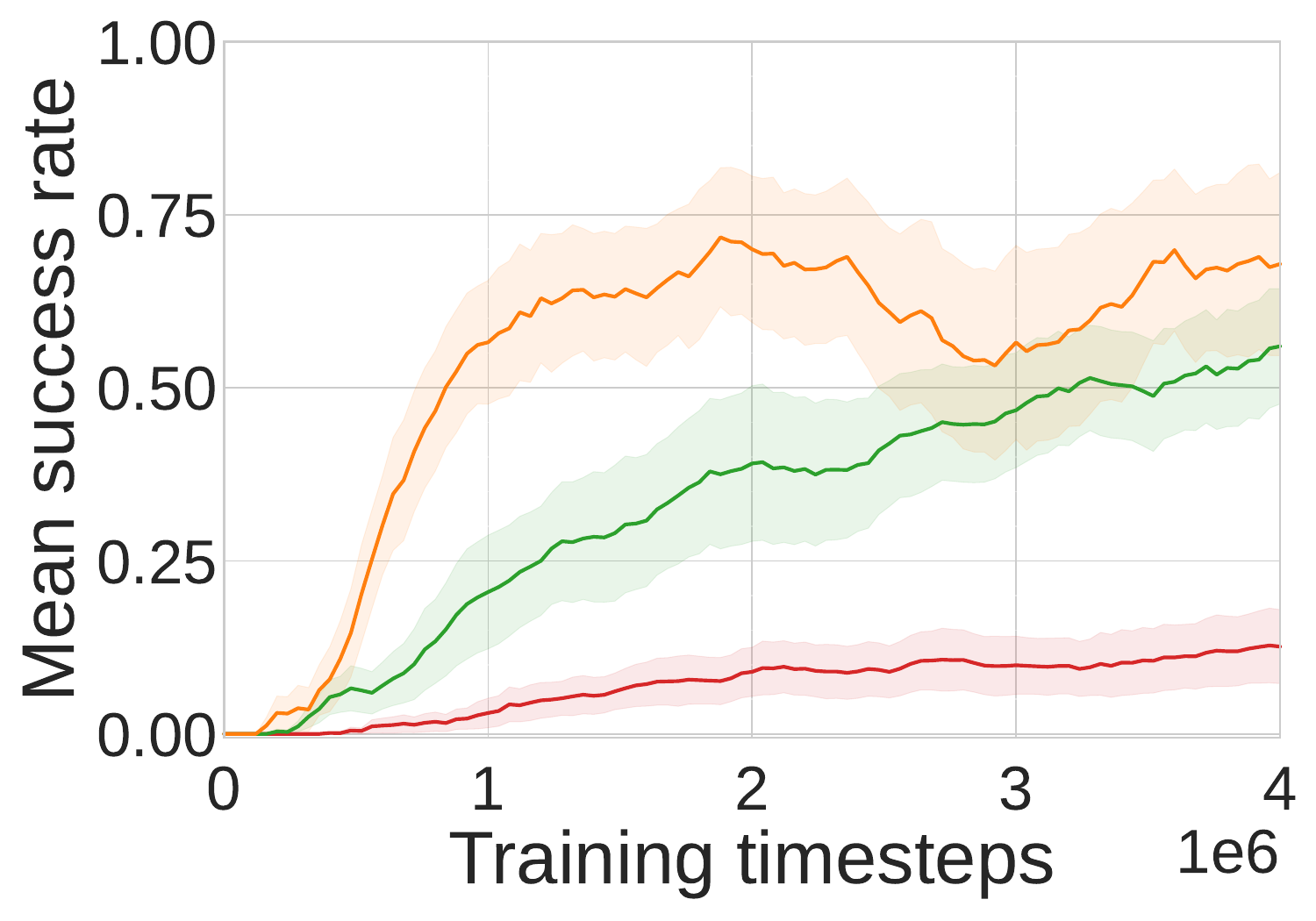} & \includegraphics[align=c, scale=0.17]{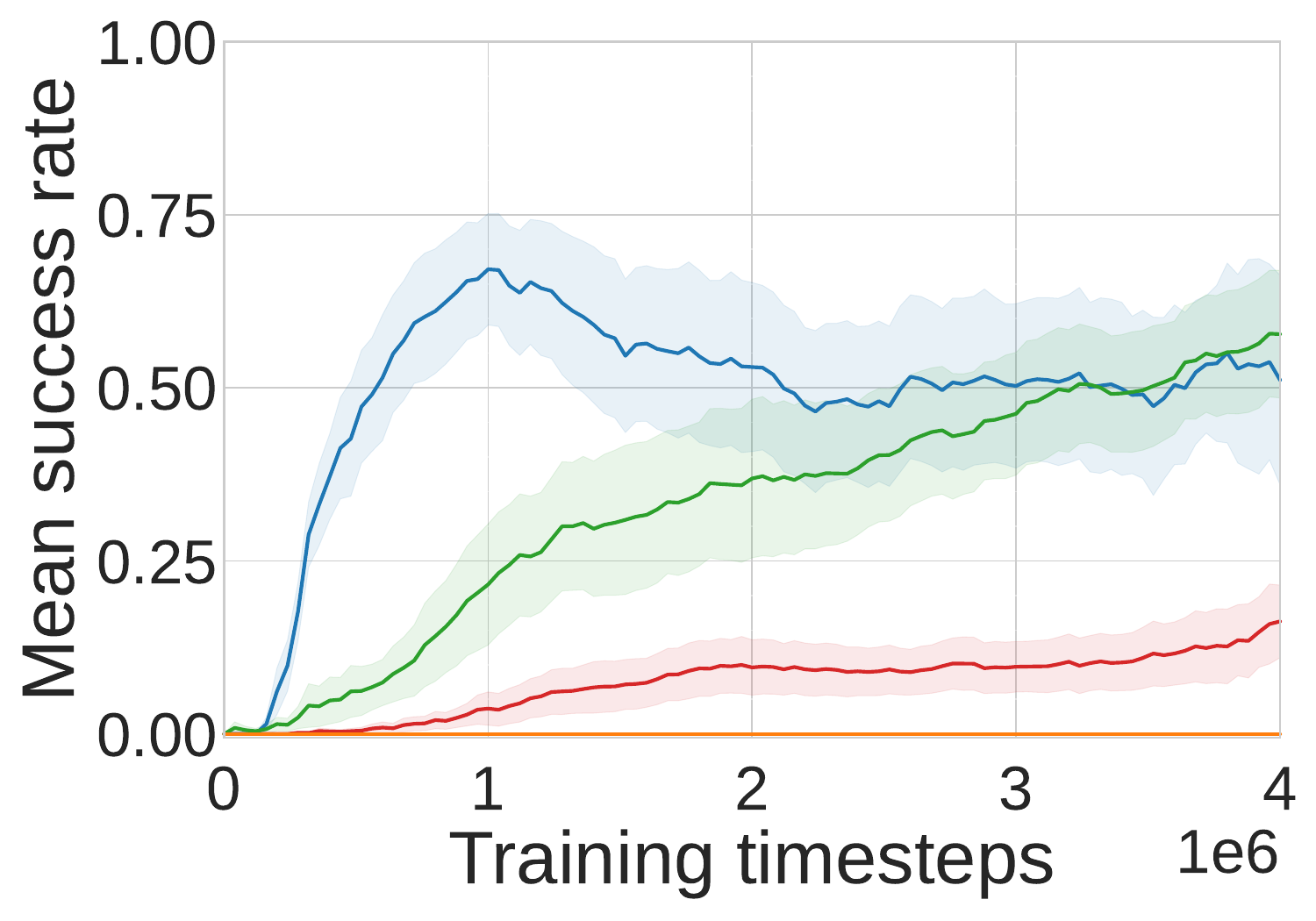} & 		\includegraphics[align=c, scale=0.17]{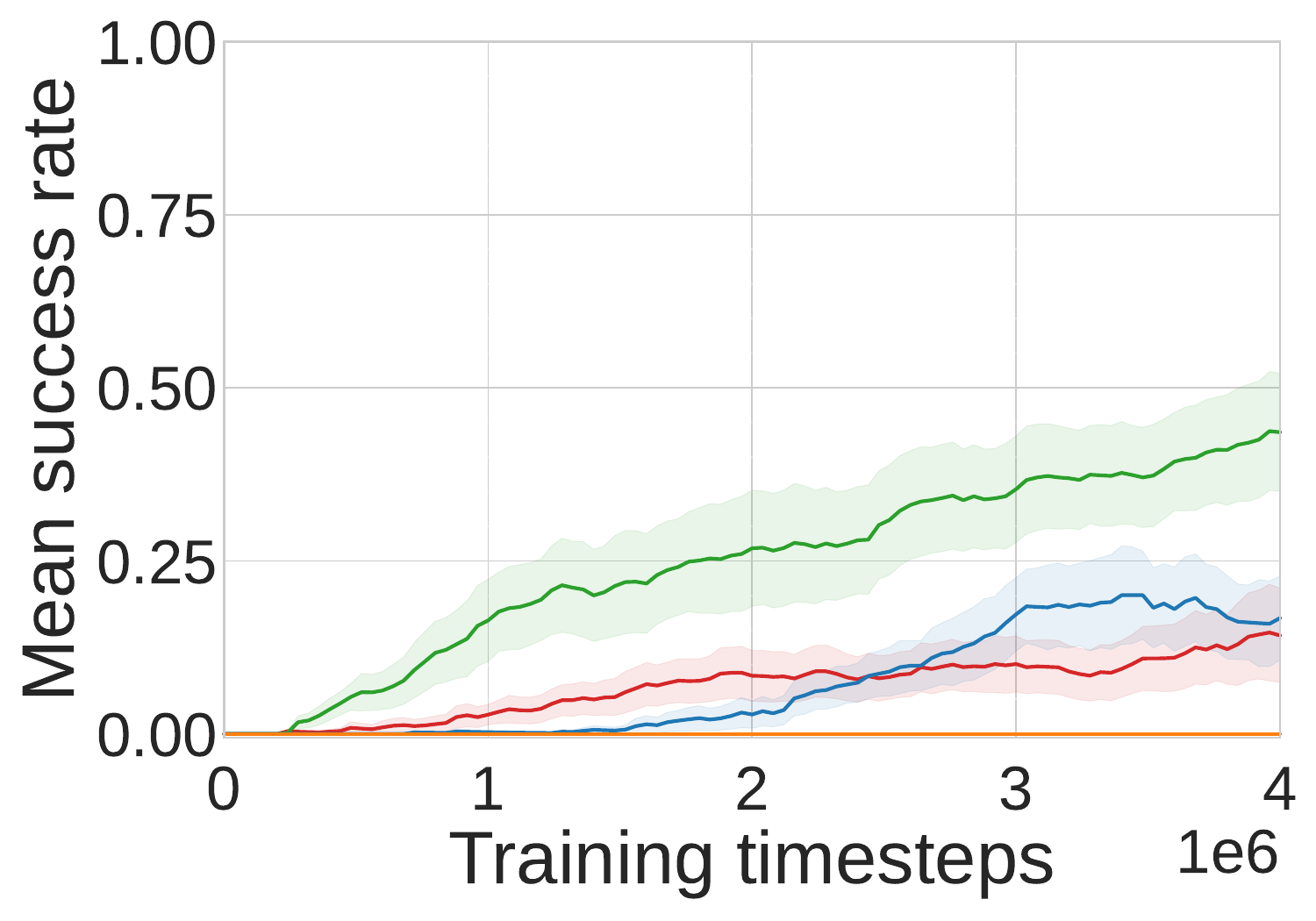} \\
   \rotatebox[origin=c]{90}{Peg Insert} & \includegraphics[align=c, scale=0.17]{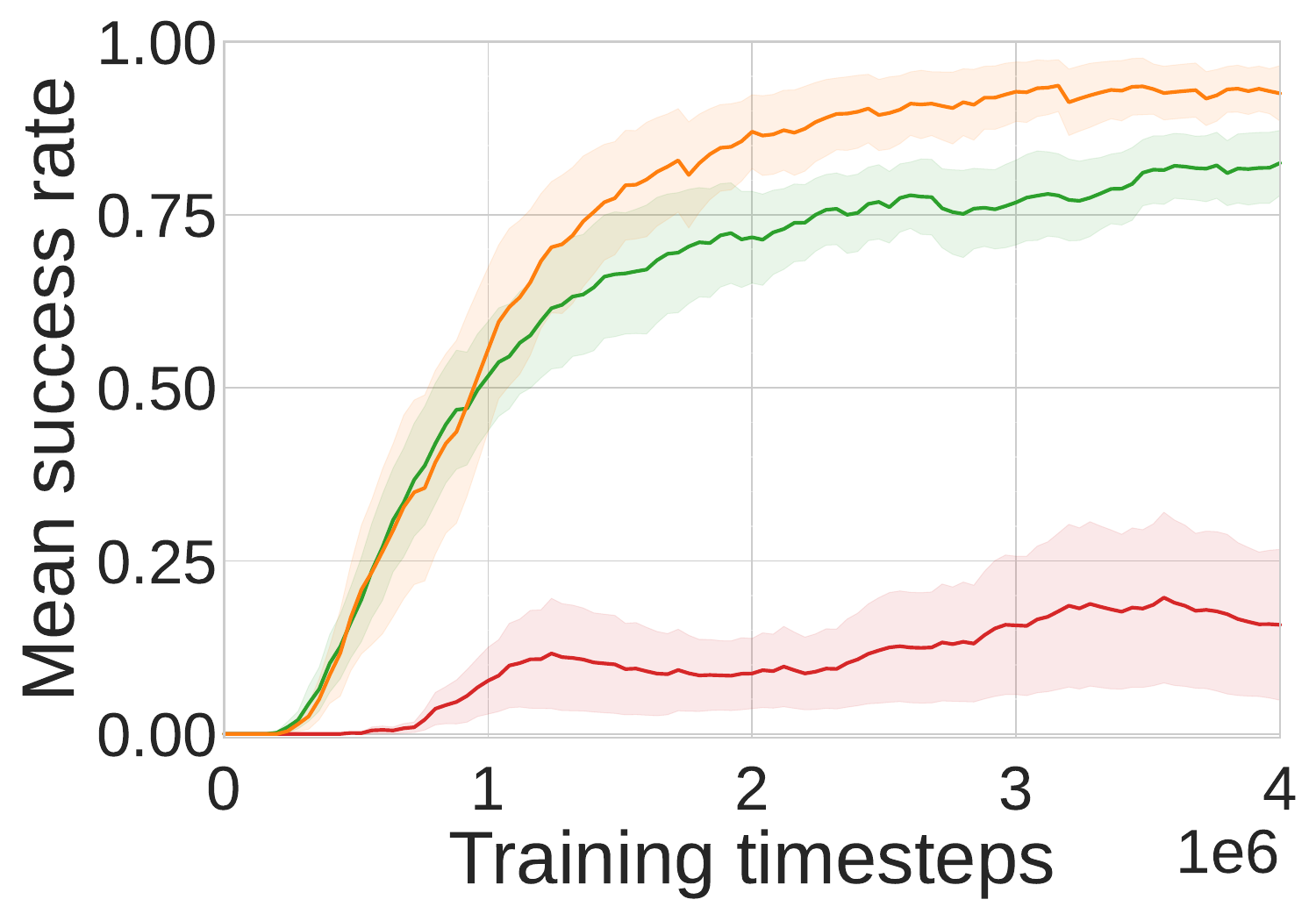} & \includegraphics[align=c, scale=0.17]{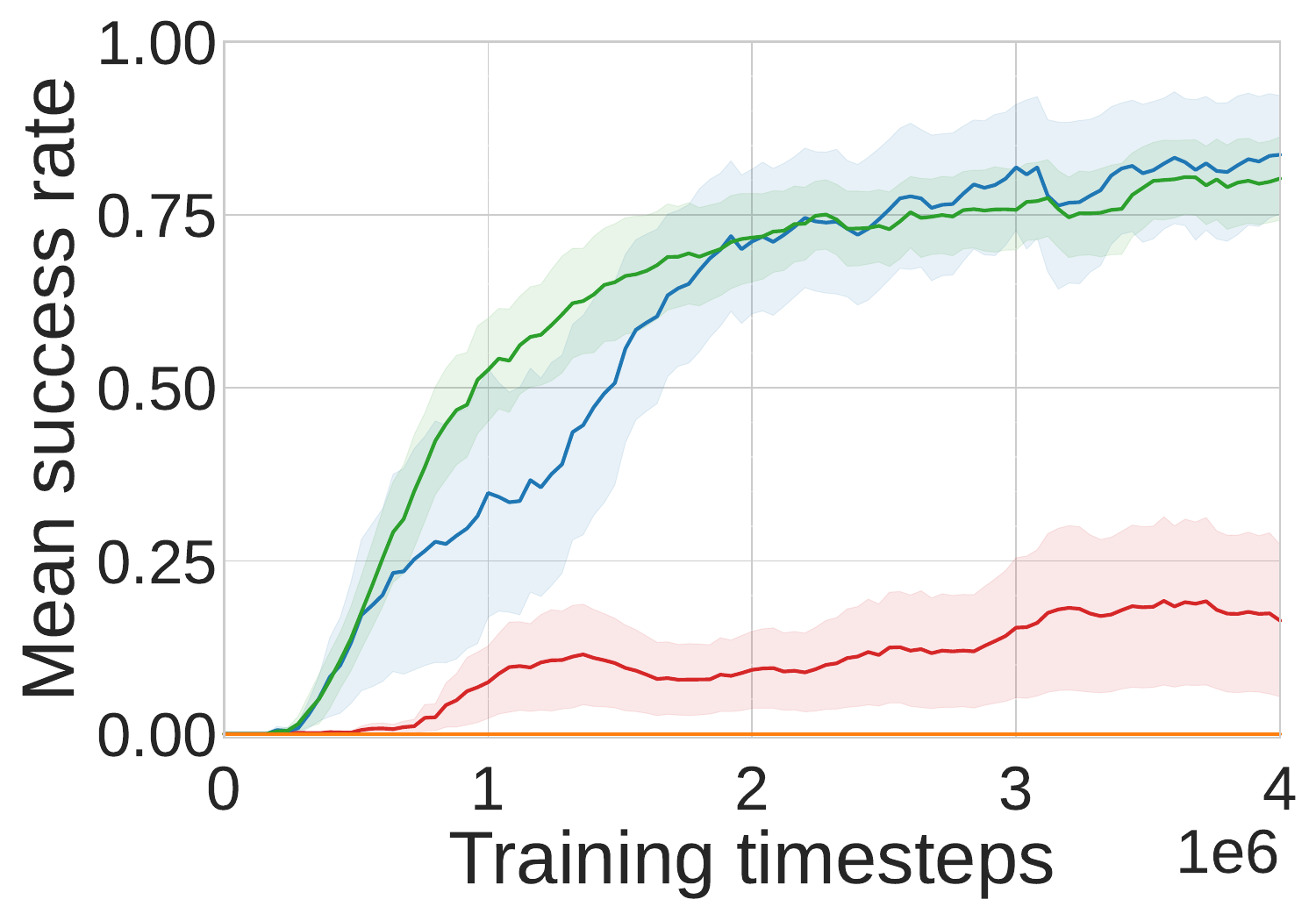} & 		\includegraphics[align=c, scale=0.17]{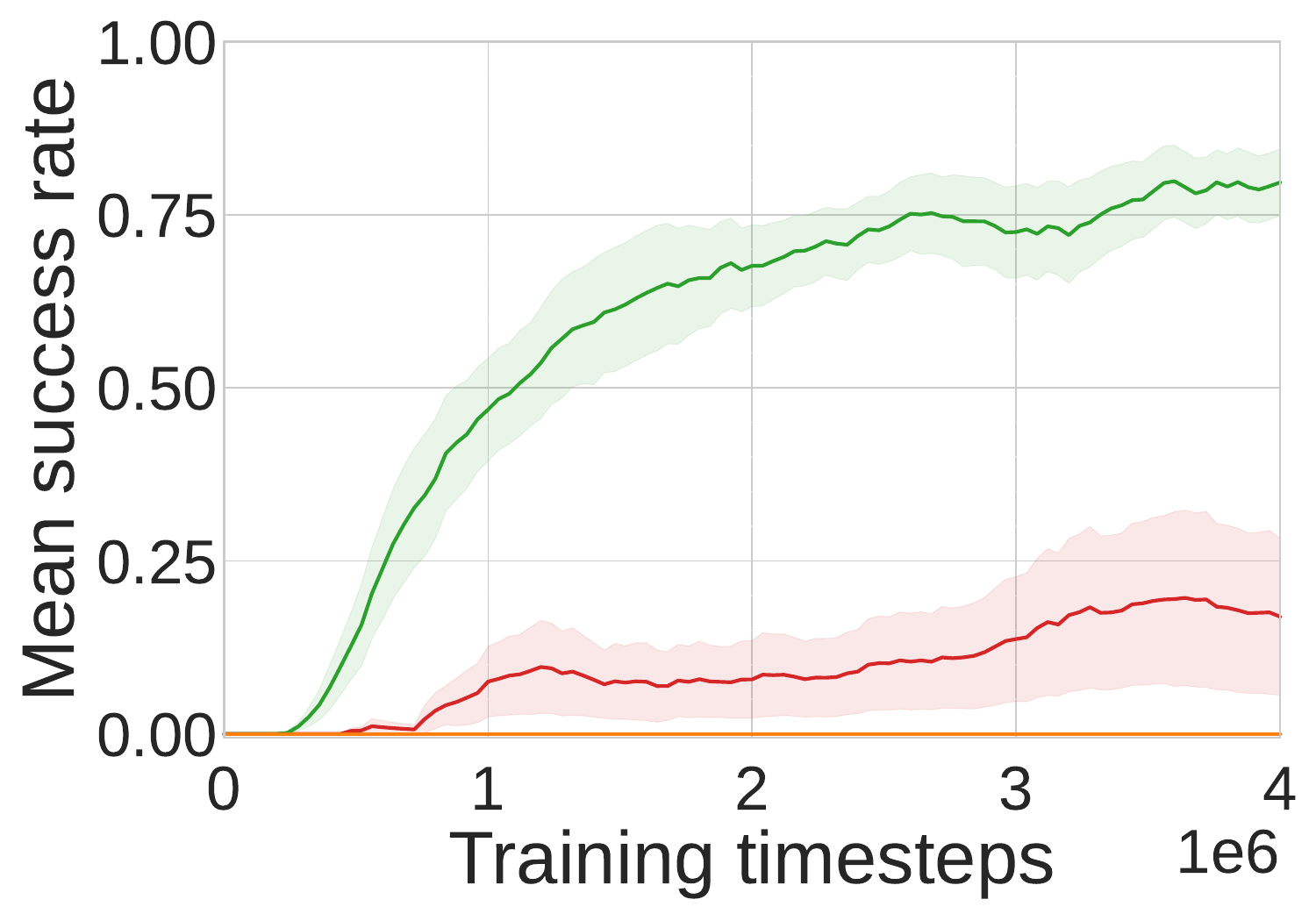} \\
\end{tabular}
\end{subfigure}
\begin{subfigure}{1.0\textwidth}
\centering
\vspace{2mm}
\includegraphics[scale=0.25]{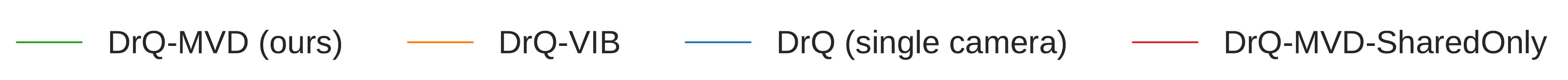}
\end{subfigure}
\caption{Results for MetaWorld tasks showing success rate for evaluation on all cameras (left of dashed line) compared with success rate on each of the individual cameras (right). Success rate is averaged over 20 evaluation episodes for 5 seeds. The shaded region is standard deviation.}
\label{fig:metaworld_results}
\end{figure}

\subsection{Generalisation results}
\label{subsec:results}
The results for Panda tasks in Figure~\ref{fig:panda_results} and MetaWorld in Figure~\ref{fig:metaworld_results}, show the task success rate of evaluation episodes completed at intervals throughout training. For MVD and VIB, which both train on all cameras, the success rate for evaluation episodes on all cameras is provided on the left of the vertical dashed line for each task. The success rate for each individual camera in the training set is provided on the right of the dashed line, representing the zero-shot generalisation performance of MVD and VIB while also showing the performance of the base RL algorithm that learns only with that individual camera.

As expected, the VIB baseline achieves optimal performance when evaluated on all three cameras but is unable to generalise to any individual camera. The base RL algorithm trained directly on each single camera is able to learn an optimal policy only for the first-person camera but fails to learn on the third-person cameras. In contrast, MVD achieves similar performance to VIB when evaluated on all cameras, albeit with lower sample efficiency in some tasks, but also achieves zero-shot generalisation to each camera, even when the base RL algorithm is unable to learn directly from that camera alone. MVD is the only method to attain consistent performance when evaluated on each camera individually. 

The ablation of MVD with the shared representation only (MVD-SharedOnly) demonstrates the importance of the private representation in MVD. In all the tasks except Panda Reach, the shared representation alone achieves lower performance than MVD. This may be because the first-person camera is easier to learn from than the third-person cameras (as evidenced by the single camera baseline). Encouraging the shared representations to be similar for all cameras, without the flexibility of a separate private representation, may prevent important features from a single camera being used during training if the agent is not yet able to extract similar features from the other cameras.

\subsection{Ablation study}
In this section, we conduct more detailed analysis of MVD for the Panda Reach and MetaWorld Soccer tasks with ablation experiments to better understand the components of MVD. We also provide some analysis of the learned representations in Appendix \ref{appendix:saliency}.

\begin{figure}
\centering
\begin{subfigure}{1.0\textwidth}
\begin{tabular}{cc:ccc}
    & all cameras & first-person & third-person (front) & third-person (side) \\
    \rotatebox[origin=c]{90}{Reach} & \includegraphics[align=c, scale=0.17]{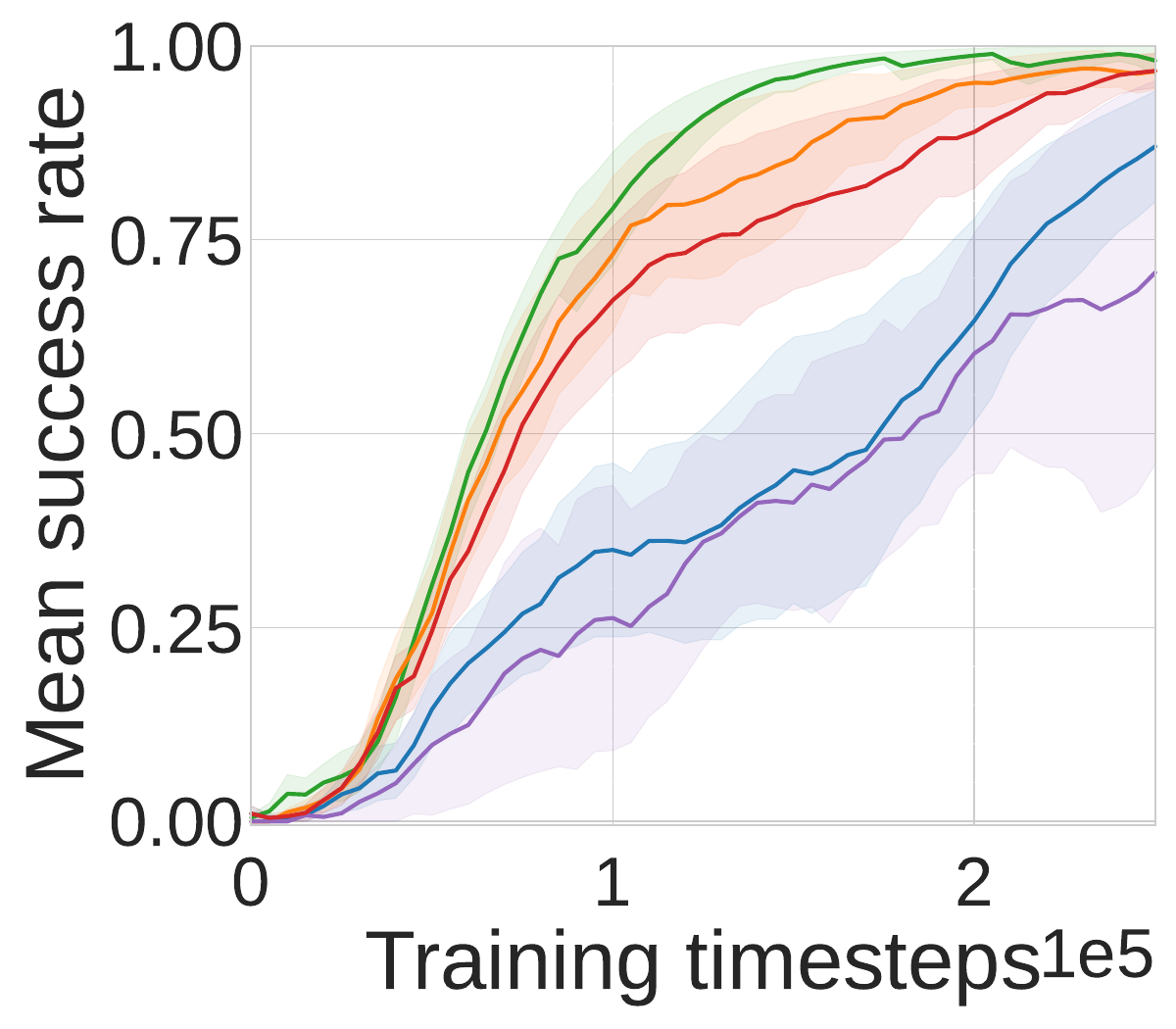} & \includegraphics[align=c, scale=0.16]{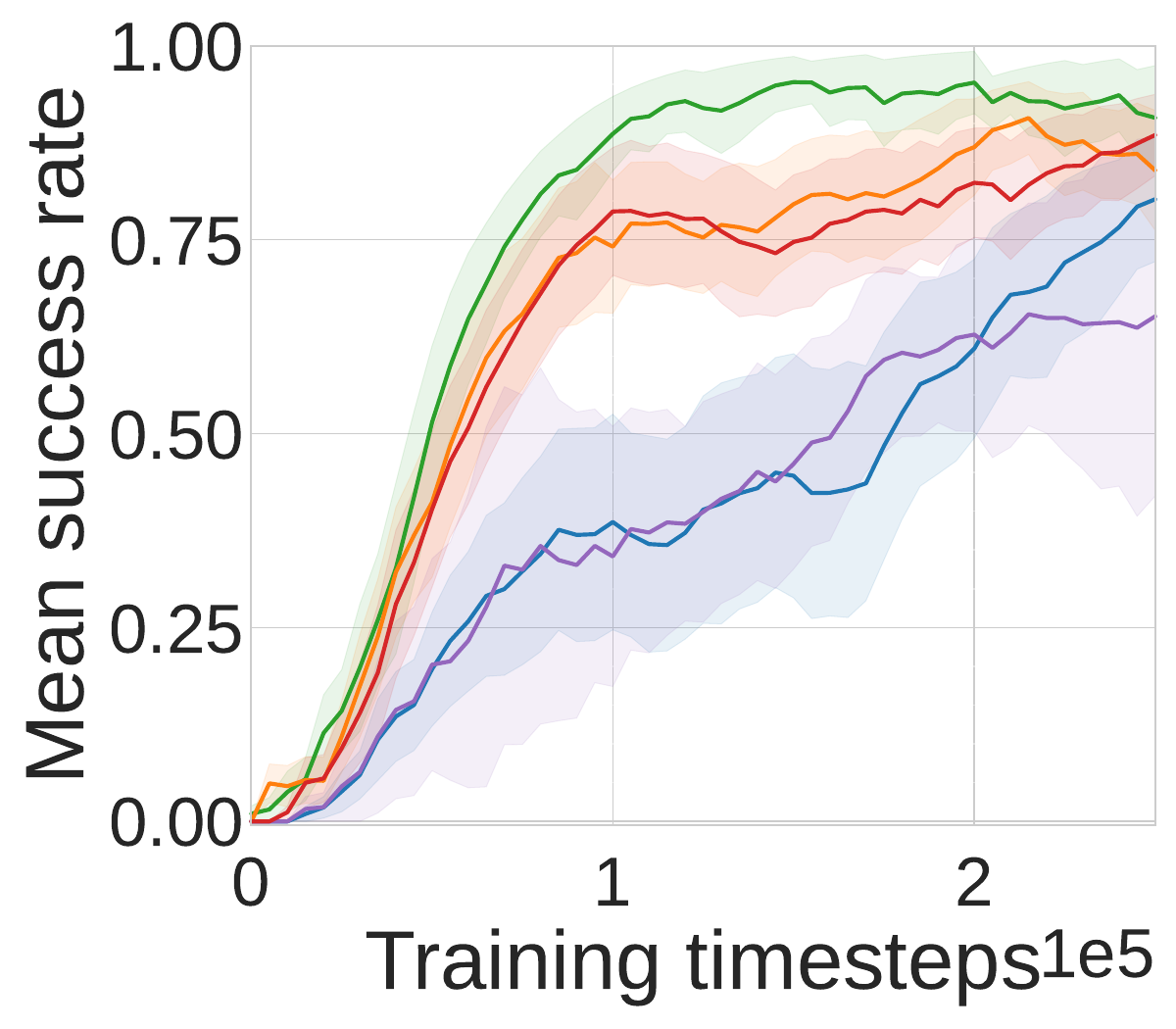} & 		
    \includegraphics[align=c, scale=0.16]{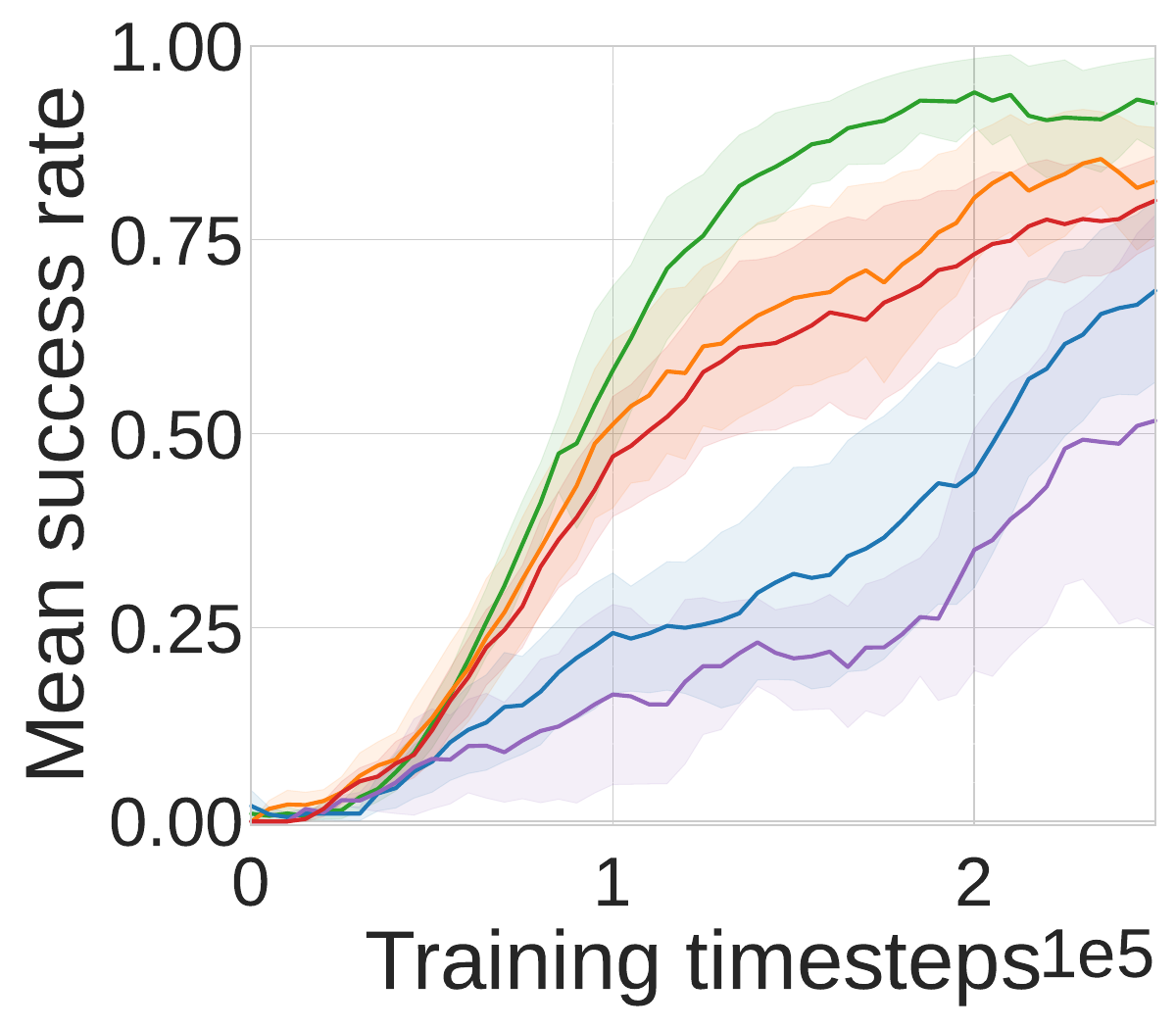} &
    \includegraphics[align=c, scale=0.16]{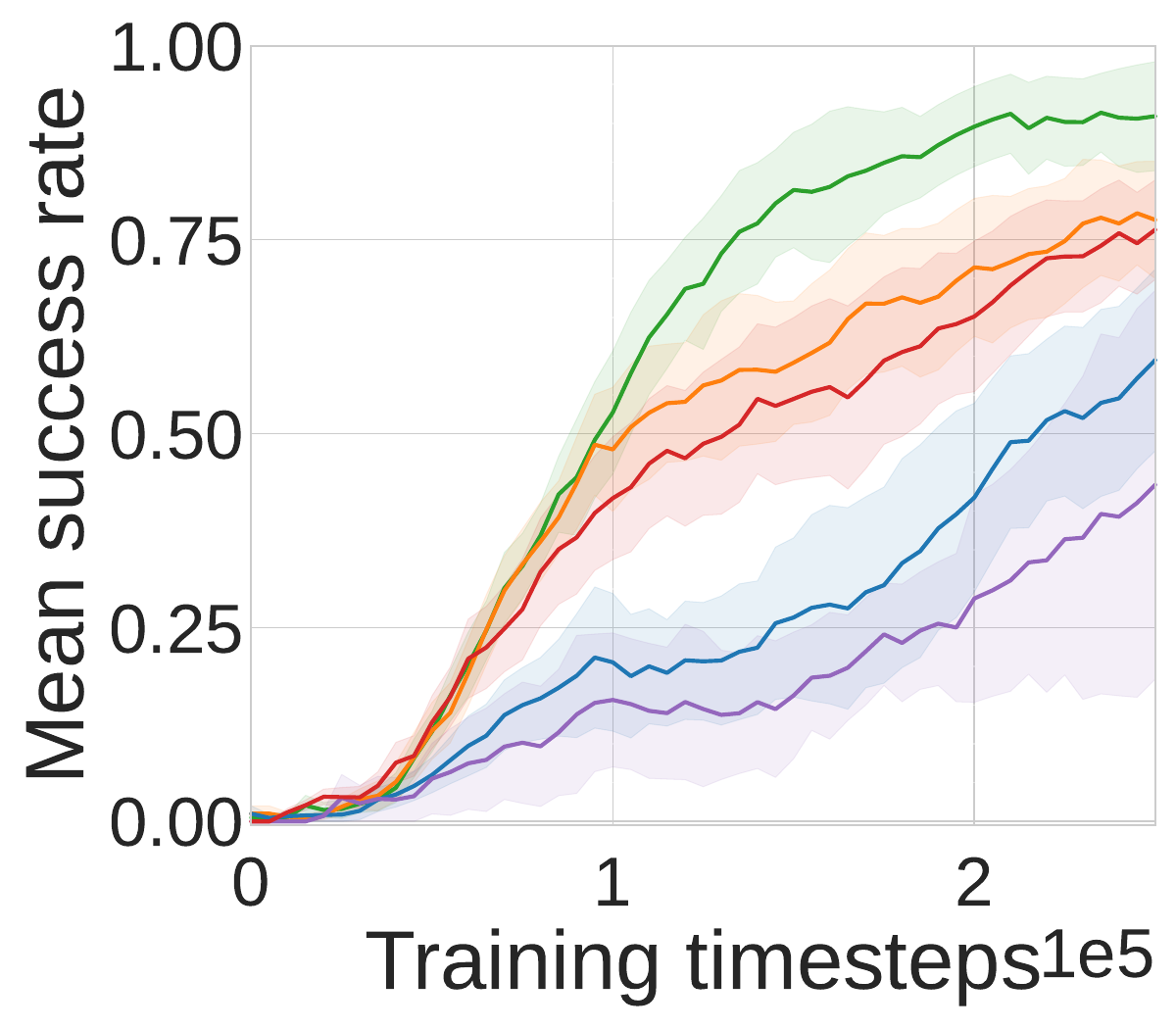} 
\end{tabular}
\end{subfigure}
\begin{subfigure}{1.0\textwidth}
\centering
\vspace{2mm}
\includegraphics[scale=0.23]{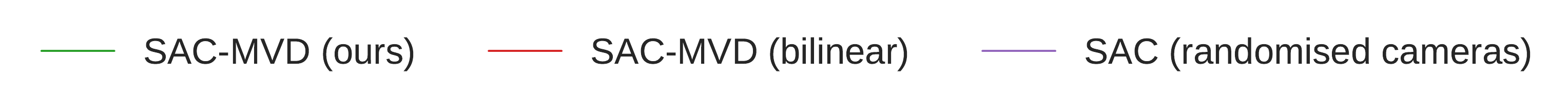}\\
\vspace{-2mm}
\includegraphics[scale=0.23]{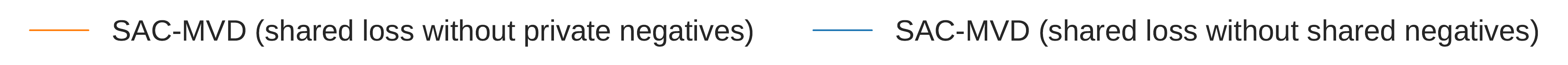}
\end{subfigure}
\caption{Results of ablation experiments for Panda Reach showing success rate for evaluation on all cameras (left of dashed line) compared with success rate on each of the individual cameras (right). Success rate is averaged over 20 evaluation episodes for 5 seeds. The shaded region is standard deviation. See text for details of each ablation.}
\label{fig:panda_ablations}
\end{figure}

\begin{figure}[t]
\begin{subfigure}{1.0\textwidth}
\centering
\begin{tabular}{cc:cc}
    & all cameras & first-person & third-person \\
    \rotatebox[origin=c]{90}{Soccer} & \includegraphics[align=c, scale=0.17]{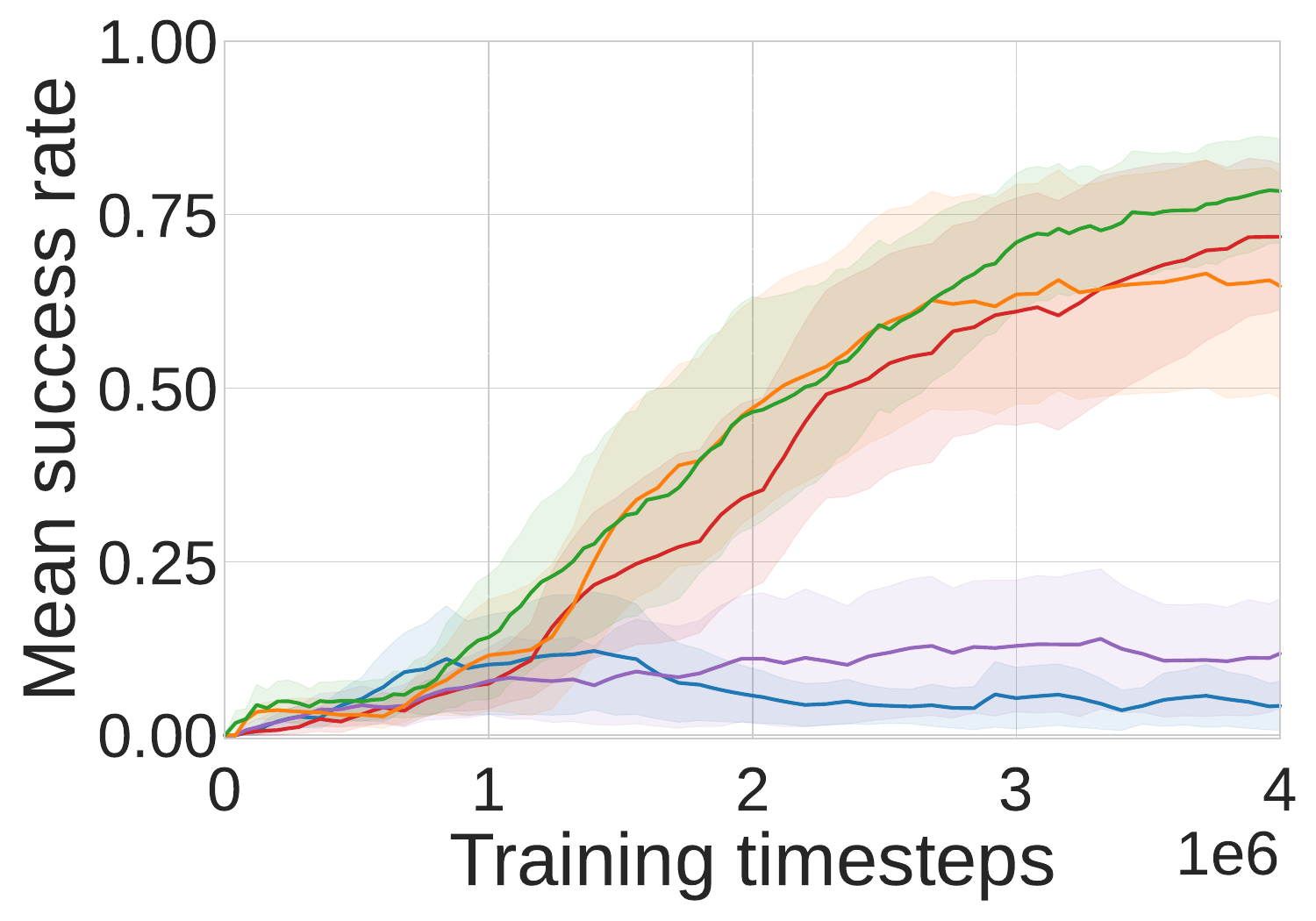} & \includegraphics[align=c, scale=0.17]{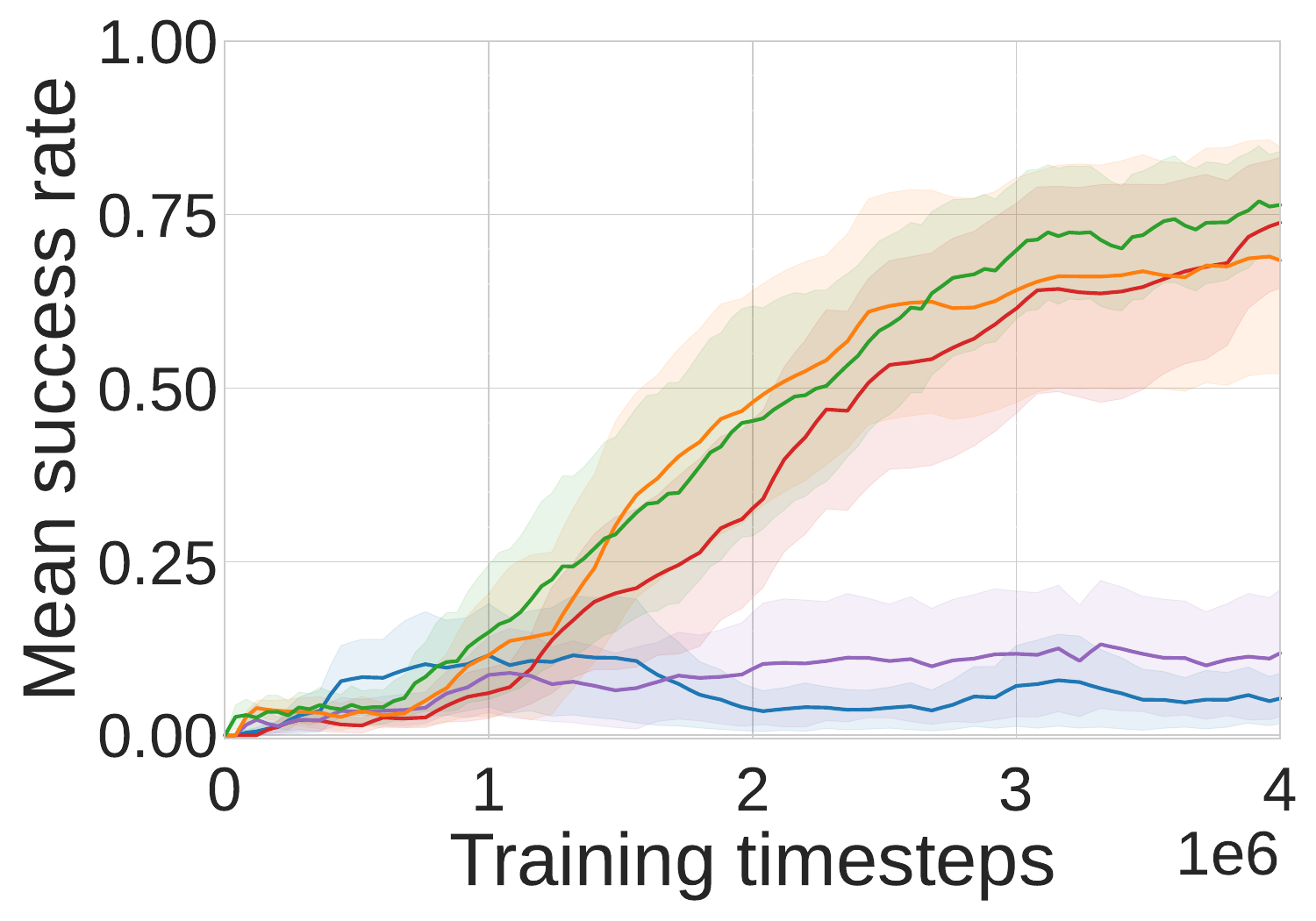} & 		\includegraphics[align=c, scale=0.17]{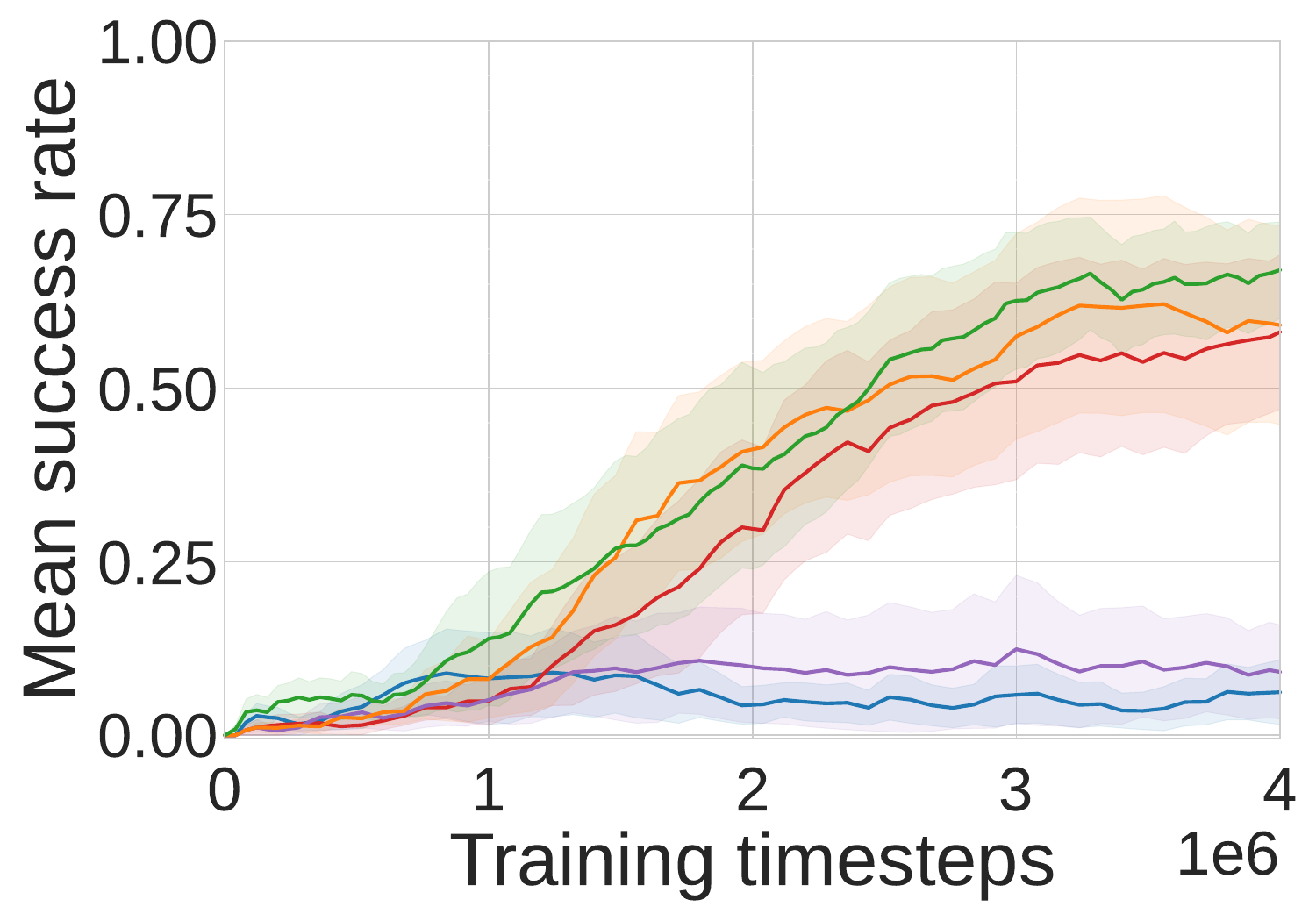}
\end{tabular}
\end{subfigure}
\begin{subfigure}{1.0\textwidth}
\centering
\vspace{2mm}
\includegraphics[scale=0.23]{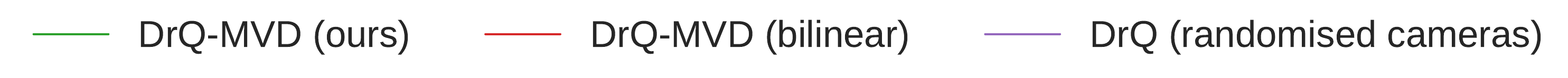}\\
\vspace{-2mm}
\includegraphics[scale=0.23]{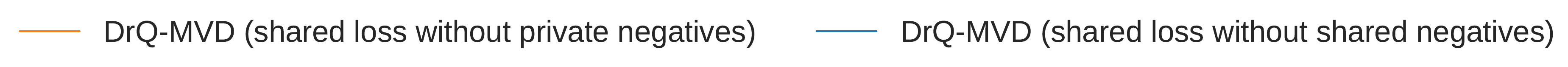}
\end{subfigure}
\caption{Results of ablation experiments for MetaWorld Soccer showing success rate for evaluation on all cameras (left of dashed line) compared with success rate on each of the individual cameras (right). Success rate is averaged over 20 evaluation episodes for 5 seeds. The shaded region is standard deviation. See text for details of each ablation.}
\label{fig:metaworld_ablations}
\end{figure}

We consider the MVD shared loss (Equation \ref{eq:mvd_shared}), which uses two types of negative samples: shared representations from the same camera at different timesteps, and private representations for all cameras at the same timestep. We assess the impact of each type of negative sample by measuring performance when one type is removed, resulting in two MVD ablations: \textit{MVD shared loss without shared negatives} and \textit{MVD shared loss without private negatives}. The results in Figures \ref{fig:panda_ablations} and \ref{fig:metaworld_ablations} show that  removing either of the negative samples reduces the performance of MVD, but the shared negatives have a much greater impact on performance than the private negatives. This may be due to the other constraints in the MVD shared and private losses indirectly encouraging disentanglement by encouraging shared representations to be similar and private representations to be dissimilar, limiting the additional improvement gained by explicitly doing so through the use of private negatives in the shared loss.

To understand the impact of the similarity measure for InfoNCE, we compare the normalised dot product similarity measure used for MVD to an alternative \textit{bilinear} similarity measure. The results, also in Figures \ref{fig:panda_ablations} and \ref{fig:metaworld_ablations}, show that while the method is reasonably robust to the choice of similarity measure, the dot product outperforms the bilinear similarity measure, particularly in Panda Reach. This may be due to the learnable weight matrix in the bilinear similarity measure making it a more lenient constraint as the agent can learn to give less weight to some dimensions than others.

Finally, we consider the same training setup as MVD but without the MVD loss. We train the base algorithm with \textit{randomised cameras} to replicate the MVD training setup by randomly sampling a camera representation as input to the RL loss at each timestep (as described in Section \ref{subsec:rl}) but without the MVD loss to structure the learned representation. As in the previous section, the base algorithm is SAC for Panda Reach and DrQ for MetaWorld Soccer. The results in Figures \ref{fig:panda_ablations} and \ref{fig:metaworld_ablations} show that randomising the cameras alone is not enough to learn to solve the more difficult MetaWorld Soccer task. While camera randomisation does improve on the single camera baseline (see Figure \ref{fig:panda_results}) for the easier Panda Reach task, it achieves a much lower success rate than MVD.

\section{Conclusion and future work}
We demonstrated that camera perspective impacts the ability of an RL agent to learn an optimal policy. The impact can be mitigated by training with multiple cameras in simulation, but hardware constraints may prevent an RL agent from always relying on access to all of these cameras in the real-world. We propose Multi-View Disentanglement (MVD), an auxiliary task for RL algorithms to learn disentangled representations with a shared representation, which is aligned across all cameras, and a private representation, which is camera-specific. Our experiments showed that an RL agent trained only on a single third-person camera cannot learn an optimal policy in many control tasks, whereas MVD, benefiting from multiple cameras in training, achieves robustness to a reduction in cameras to solve the task with the same single third-person camera.

Future work could leverage progress on feature-level disentanglement, either within RL \citep{Dunion2023cmid, Dunion2023ted, Higgins2017darla} or in the multi-view disentanglement literature \citep{Qui2023, Hsieh2018LearningTD}, to further extend this work to disentangle individual features within the shared and private representations. Future work could also consider extending to a larger number of cameras to learn a representation that generalises out-of-distribution for sim2real transfer when real-world camera views do not perfectly match simulation.

\bibliography{main}
\bibliographystyle{rlc}

\clearpage
\appendix

\section{Extended background}
\label{appendix:background}
We use SAC~\citep{Haarnoja2018sac} and DrQ \citep{Yarats2021drq} as the base RL algorithms
for the MVD auxiliary task.

SAC is an actor-critic RL algorithm for continuous control. SAC maximises the expected return and entropy of the policy $\pi$. The critic $Q$ minimises the loss:
\begin{equation}
	L_{Q} = \mathbb{E}_{(\mathbf{o}_t, \mathbf{a}_t, \mathbf{o}_{t+1}, r_t) \sim \mathcal{D}}\left[ \left(Q(\mathbf{o}_t,\mathbf{a}_t) - r_t - \gamma \bar{V}(\mathbf{o}_{t+1}))\right)^2 \right]
\end{equation}
where $\mathbf{o}_t$ is the image observation and $\mathbf{a}_t$ is the action at time $t$. The actor $\pi$ is trained by minimising the loss:
\begin{equation}\label{eqn:policy_loss}
	L_{\pi} = \mathbb{E}_{\mathbf{o}_t \sim \mathcal{D}} \left[ \mathbb{E}_{\mathbf{a}_t \sim \pi} \left[ {\alpha}_{\text{SAC}} \log (\pi(\mathbf{a}_t \mid \mathbf{o}_t)) - \min_{i=1,2}\bar{Q}_i(\mathbf{o}_t,\mathbf{a}_t) \right] \right]
\end{equation}
where $\bar{Q}$ is exponential moving average of the Q network parameters. We augment SAC with a decoder, trained with an image reconstruction loss, to improve learning from images. The encoder and decoder details are provided in Appendix~\ref{appendix:implementation}.

DrQ is a data augmentation approach for robust learning from image pixels without the need for a decoder. DrQ adds padding and random crop augmentations to the image observations and averages over the target Q-value for each augmentation in the critic update as well as averaging over the augmentations for the Q function itself. The actor $\pi$ uses unaugmented images and applies the SAC policy loss in Equation~\ref{eqn:policy_loss}.

\section{Implementation details}
\subsection{MVD implementation}
\label{appendix:implementation}
Our codebase is built on top of the public and open-source generalisation benchmark code provided by \cite{hansen2021softda}, and uses the official DrQ implementation by~\citet{Yarats2021drq}. A public and open-source implementation of MVD is available at \href{https://github.com/uoe-agents/MVD}{github.com/uoe-agents/MVD}.

\paragraph{Encoders.} Both the shared encoder $f_{\theta}$ and private encoder $g_{\phi}$ consist of 4 convolutional layers, each with a $3\times3$ kernel size and 32 channels. The first layer has a stride of 2, all other layers have a stride of 1. There is a $\text{ReLU}$ activation between each of the convolutional layers. The convolutional layers are followed by a linear layer, normalisation, then a tanh activation function. The output size (i.e. size of each representation) is 50. The encoder weights are shared between the actor $\pi$ and critic $Q$.

\paragraph{Decoder.} Where we use SAC as the base RL algorithm, we also include a decoder trained with an image reconstruction loss as this improves the sample efficiency of SAC with images (e.g. \cite{Yarats2021sacae}). The decoder is not used for DrQ. The first layer of the decoder is a fully-connected layer, which is followed by 4 deconvolutional layers, each with a $3\times3$ kernel size and 32 channels. Each deconvolutional layer has a stride of 1, except the last, which has a stride of 2. The reconstruction loss, which is used to update both the encoder and the decoder, is the mean squared error between the input image and the reconstructed image.

\paragraph{Actor and critic.}
Both the actor $\pi$ and critic $Q$ networks are multilayer perceptrons that each consist of two layers and have a hidden dimension of 1024. There is a $\text{ReLU}$ activation after each layer except the last layer.

\paragraph{Hyperparameters.} Table~\ref{table:hyperparams} shows the hyperparameters for all tasks for both MVD and baselines as they use the same base RL algorithm.

\begin{table}[!ht]
	\centering
	\begin{tabular}{ | c | c | }
		\hline
		Hyperparameter & Value \\ 
		\hline 
		Replay buffer capacity & 100000 \\
		Initial steps before training begins & 1000 \\
		Stacked frames & 3 for MetaWorld, 1 for Panda \\
		Action repeat & 2 for MetaWorld, 1 for Panda \\ 
		Batch size & 128 \\
		Discount factor & 0.99 \\
		Optimizer & Adam \\
		Learning rate (actor, critic and encoder) & 1e-3 \\
            $\alpha_{\text{SAC}}$ learning rate & 1e-4 \\
		Q function soft-update rate & 0.01 \\
		Actor update frequency & 2 \\
		Actor log stddev bounds & $[-10,2]$ \\
		Initial temperature & 0.1 \\
            Image size & $ 3 \times 84 \times 84$ \\
            InfoNCE temperature & 0.1 \\
		\hline 
	\end{tabular}
	\vspace{1ex}
	\caption{Hyperparameter values.}
\label{table:hyperparams}
\end{table}

\section{Representation analysis}
\label{appendix:saliency}
We use saliency maps to show that our learned representations reflect shared and private features in the camera images. The attribution method we use is Integrated Gradients \citep{Sundararajan2017ig} with SmoothGrad-Squared \citep{Hooker2018ABF} to reduce visual noise. However, while features attribution methods such as Integrated Gradients are commonly used in the literature, very recently \cite{Bilodeau2024} has shown that such feature attributions methods can fail to improve on random guessing for inferring model behaviour. We provide our representation analysis results for completeness. 

\subsection{Saliency map results}
\begin{figure}[t]
\centering
\begin{tabular}{cll}
& \multicolumn{1}{c}{\hspace{-9mm}{Panda Reach}} 
& \multicolumn{1}{c}{\hspace{-7mm}{MetaWorld Soccer}}  
\vspace{3mm} \\
& \hspace{3mm} first-person \hspace{3mm} third-person \hspace{3mm} third-person & \hspace{2mm} first-person \hspace{4mm} third-person \\
& \hspace{31mm} (front) \hspace{12mm} (side)  \\
    \rotatebox[origin=c]{90}{Images} & \includegraphics[align=c, scale=0.455]{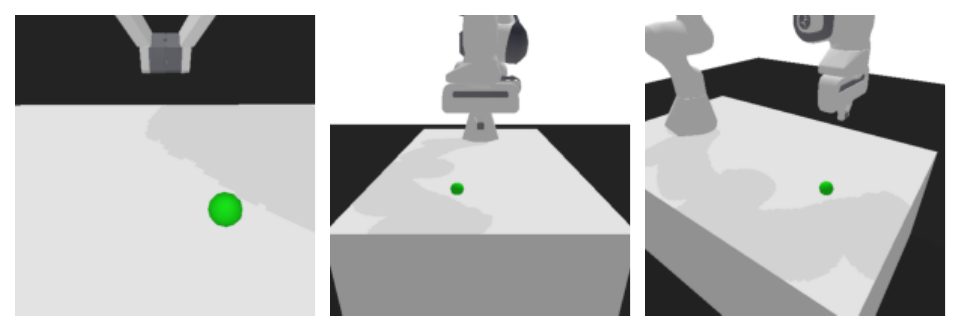} & \includegraphics[align=c, scale=0.3]{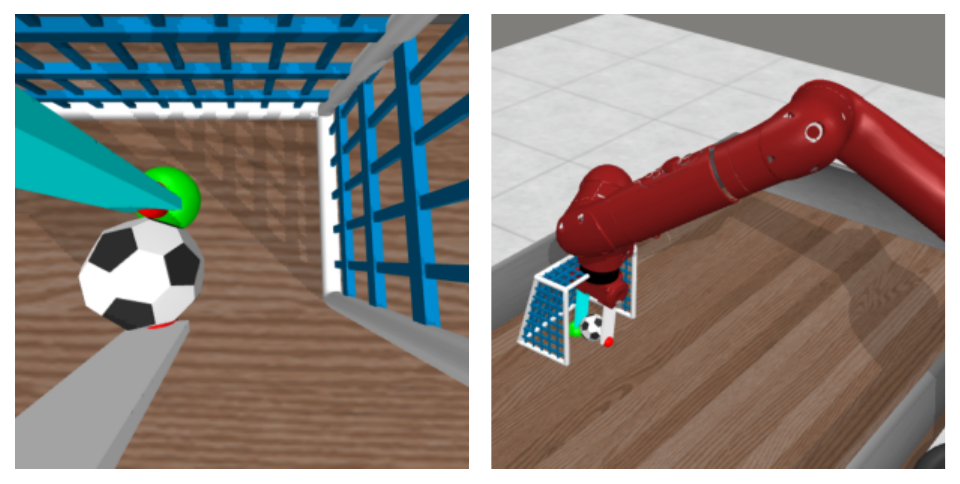} \\
    \rotatebox[origin=c]{90}{Shared} & \includegraphics[align=c, scale=0.45]{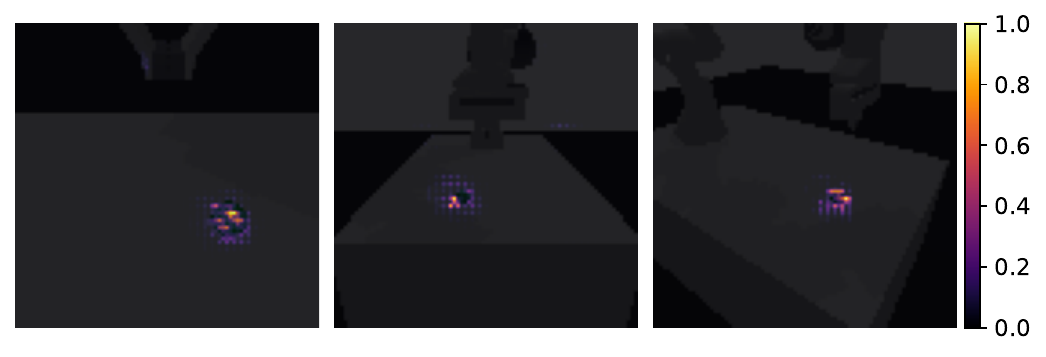} & \includegraphics[align=c, scale=0.3]{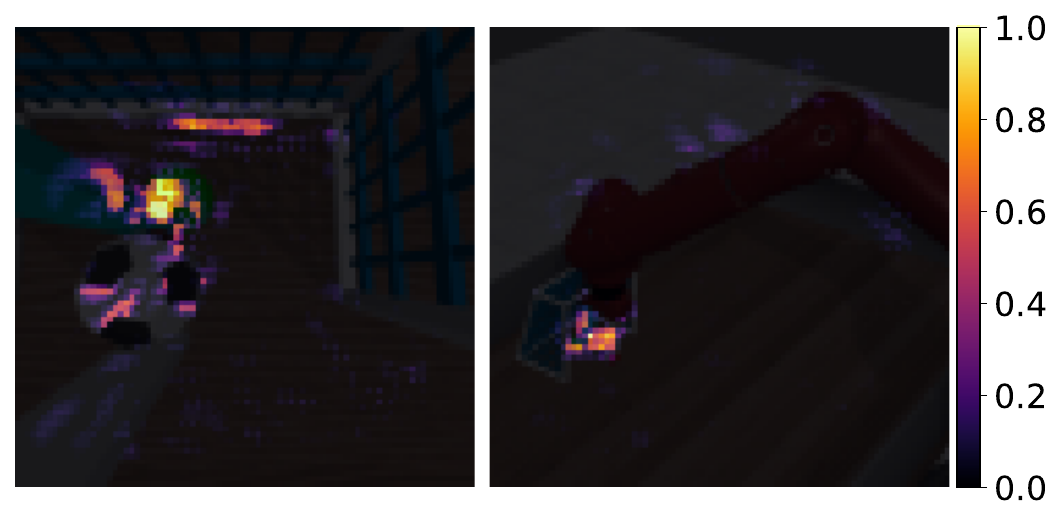} \\
    \rotatebox[origin=c]{90}{Private} & \includegraphics[align=c, scale=0.45]{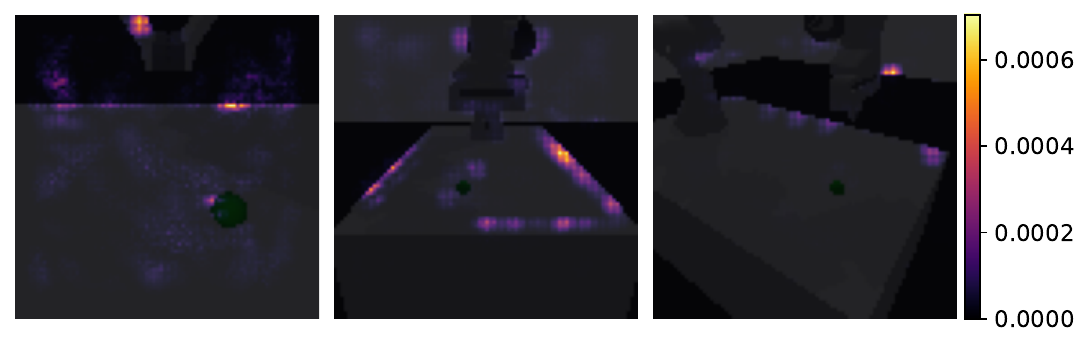} & \includegraphics[align=c, scale=0.3]{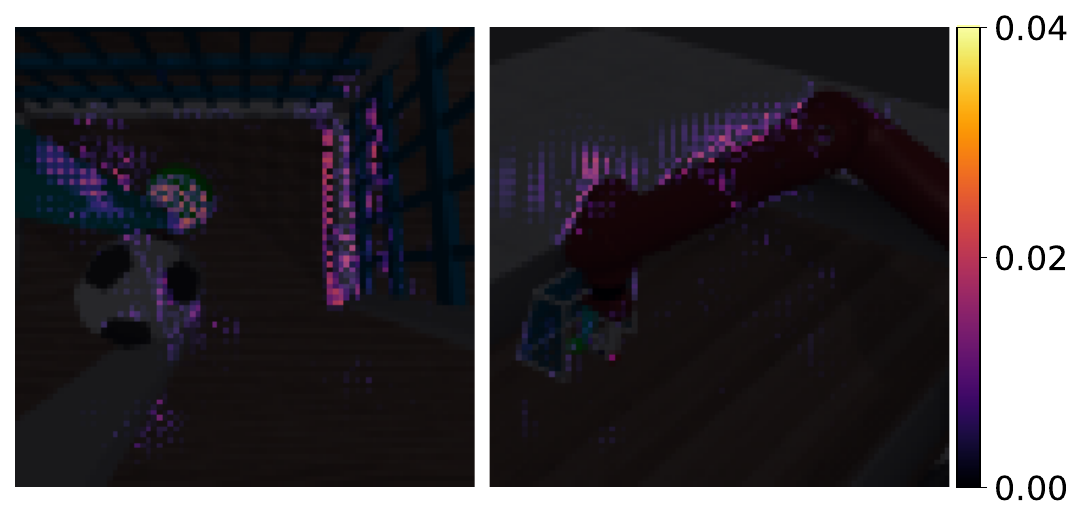}
\end{tabular}
\caption{Policy saliency maps for MVD on the Panda Reach and MetaWorld Soccer tasks. The first row shows the original images used to calculate the attributions for each camera, the second and third rows show the saliency maps for the shared and private representations, respectively. Brighter pixels correspond to higher attributions.}
\label{fig:saliency}
\end{figure}
Using the learned encoders $f_{\theta}$ and $g_{\phi}$, we calculate the attributions for each image pixel on each representation feature. We also calculate the attributions for each dimension in the representations on the output of the learned policy $\pi$. The pixel attributions are weighted by the corresponding policy attributions, and normalised for each camera to be in $[0,1]$, to visualise the attention of the RL policy based on the features in the shared and private representations separately. 

The resulting saliency maps in Figure~\ref{fig:saliency} show that the shared and private representations focus on different features. The shared representation consistently focuses on features that are visible in all cameras, such as goal position for both Panda Reach and MetaWorld Soccer. 
The private representation focuses on camera-specific features that are not clearly visible in all cameras. In Panda Reach, the private representation for the first-person camera highlights the end-effector, which may be harder to extract from the third-person cameras; while the third-person front camera highlights the table edge, which is not visible in the first-person camera. In MetaWorld Soccer, the private representation for the first-person camera highlights the right edge of the goal post, which may help the agent guide the ball into the goal but is obscured by the robot in the third-person camera. The shared and private saliency maps are shown on different scales for readability. Comparing the maximum attribution on each of the scales shows that the attributions for the shared representations are higher than the corresponding private representations. This means that the policy focuses more on the shared representation by the end of training since the most important features for the task are visible in all cameras, such as the goal positions. However, the private representation is still required during training as evidenced by the ablation experiment in Section~\ref{subsec:results}.

\subsection{Implementation details}
We use the Integrated Gradients implementation from the open-source Captum library \citep{Kokhlikyan2020} to calculate the attributions for the saliency maps in Figure~\ref{fig:saliency}. We use an input image for each camera (depicted in the first row of Figure~\ref{fig:saliency}) and an all black image as the baseline. We used SmoothGrad-Squared to reduce visual noise, which adds Gaussian noise to $n$ copies of the input image, calculates the Integrated Gradient attributions for each of these noisy images, and returns the mean squared of the attributions across the noisy images. We use the implementation of SmoothGrad-Squared provided by Captum with the default $n=5$.

For the input image $\mathbf{o}_t^{c_i}$ for each camera $c_i \in C$, we calculated the image pixel attributions for each dimension of the output for both the shared encoder $f_{\theta}$ and the private encoder $g_{\phi}$. For the corresponding representation $\mathbf{z}_t^{c_i} = \left(\mathbf{s}_t^{c_i}, \mathbf{p}_t^{c_i} \right)$, we calculated the attribution of each dimension of both the shared and private representation on the output of the policy network $\pi(\mathbf{a}_t | \mathbf{z}_t)$. The pixel attributions for each dimension of the shared and private representation are weighted by the corresponding attributions from the policy network. 
The absolute value of the attributions are summed over all dimensions for each of the shared and private representations. The resulting pixel attributions allow us to visualise the attention of the policy network to the pixels based on whether those pixels were used by the shared or private representation. The attributions are re-scaled be in $[0,1]$ for each camera for easier comparison across cameras and different representations. These normalised attributions are overlayed onto the input image to create the saliency maps. The shared and private saliency maps are shown on a different scale to improve readability of the private attributions since they are much smaller than the shared attributions.  

\section{Environment details}
In this section, we provide a description and images of each task used in our experiments.

\subsection{Panda tasks}
\begin{figure}[!ht]
\centering
\begin{subfigure}[b]{0.65\textwidth}
\begin{subfigure}[b]{0.3\textwidth}
    \centering
    \includegraphics[scale=0.35]{images/panda_camera2.png}
\end{subfigure} 
\hfill
\begin{subfigure}[b]{0.3\textwidth}
    \centering
    \includegraphics[scale=0.35]{images/panda_camera0.png}
\end{subfigure} 
\hfill
\begin{subfigure}[b]{0.3\textwidth}
    \centering
    \includegraphics[scale=0.35]{images/panda_camera1.png}
\end{subfigure} 
\caption{Panda Reach}
\vspace{7mm}
\end{subfigure}
\hfill
\begin{subfigure}[b]{0.65\textwidth}
\begin{subfigure}[b]{0.3\textwidth}
    \centering
    \includegraphics[scale=0.35]{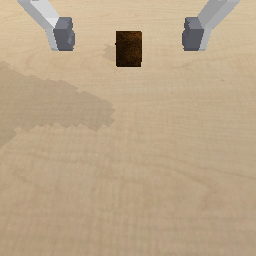}
\end{subfigure} 
\hfill
\begin{subfigure}[b]{0.3\textwidth}
    \centering
    \includegraphics[scale=0.35]{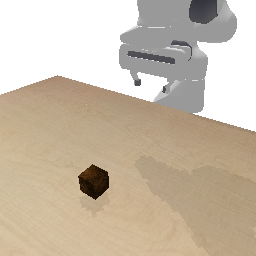}
\end{subfigure} 
\hfill
\begin{subfigure}[b]{0.3\textwidth}
    \centering
    \includegraphics[scale=0.35]{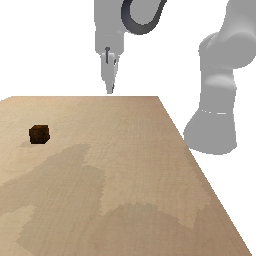}
\end{subfigure} 
\caption{Panda Cube Grasping}
\end{subfigure}
\caption{Images showing each camera view used for the Panda tasks.}
\label{fig:all_panda_cams}
\end{figure}

We evaluate performance using tasks with the simulated Franka Emika Panda robotic arm instantiated in the PyBullet physics engine \citep{coumans2019}. For both tasks, we use three camera views as the observations generated with PyBullet, depicted in Figure~\ref{fig:all_panda_cams}.

\paragraph{Panda Reach.} We use the Reach task from Panda Gym \citep{Gallouedec2021pandagym}, where the goal is for the robot to place it's end-effector at a target position. We use the dense reward setting in which the agent receives a reward at each timestep based on the distance to the goal. The action space consists of the 3D end-effector position and 1D gripper control. The position of the end-effector and goal are randomly initialised at the start of each episode.

\paragraph{Panda Cube Grasping.} The Cube Grasping task was proposed by \cite{Hsu2022VisionManipulators}. The goal is for the robot to grasp and pick up a cube. The reward at each timestep is based on the distance to the cube plus additional rewards if the agent is gripping or lifting the cube. The action space consists of the 3D end-effector position and 1D gripper control. The cube and end-effector position are randomly initialised at the start of each episode.

\subsection{MetaWorld tasks}
\begin{figure}[H]
\centering
\begin{subfigure}[b]{0.45\textwidth}
\begin{subfigure}[b]{0.45\textwidth}
    \centering
    \includegraphics[scale=0.2]{images/soccer_first_person.png}
\end{subfigure} 
\hfill
\begin{subfigure}[b]{0.45\textwidth}
    \centering
    \includegraphics[scale=0.2]{images/soccer_third_person.png}
\end{subfigure}  
\caption{Soccer}
\vspace{7mm}
\end{subfigure}
\hfill
\begin{subfigure}[b]{0.45\textwidth}
\begin{subfigure}[b]{0.45\textwidth}
    \centering
    \includegraphics[scale=0.2]{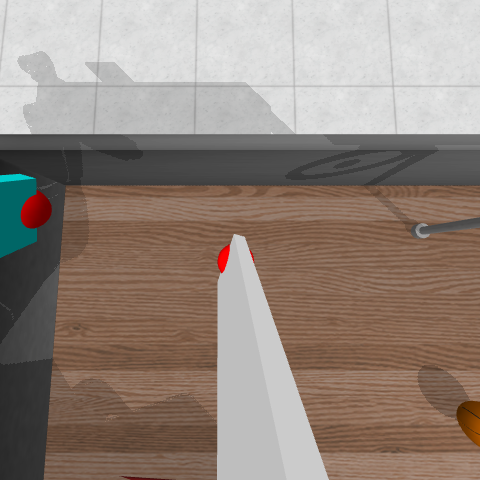}
\end{subfigure} 
\hfill
\begin{subfigure}[b]{0.45\textwidth}
    \centering
    \includegraphics[scale=0.2]{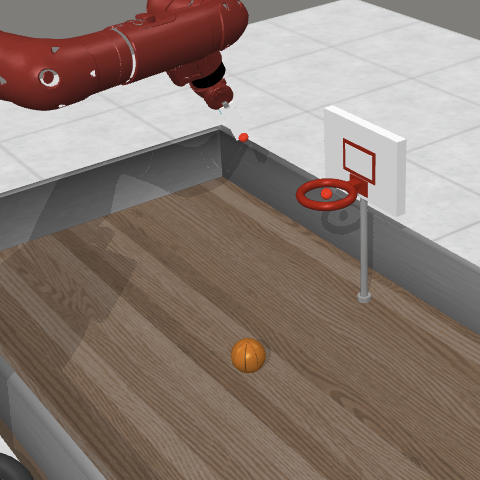}
\end{subfigure} 
\caption{Basketball}
\vspace{7mm}
\end{subfigure}
\begin{subfigure}[b]{0.45\textwidth}
\begin{subfigure}[b]{0.45\textwidth}
    \centering
    \includegraphics[scale=0.2]{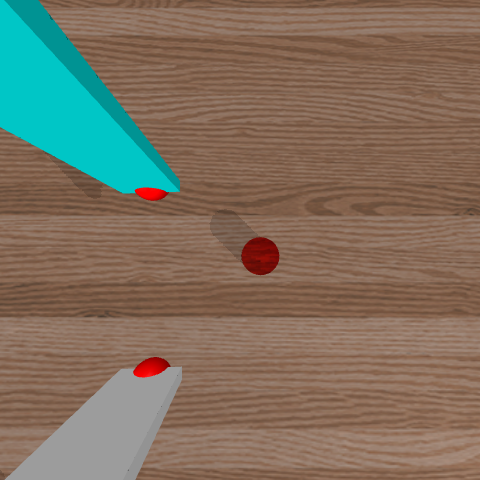}
\end{subfigure} 
\hfill
\begin{subfigure}[b]{0.45\textwidth}
    \centering
    \includegraphics[scale=0.2]{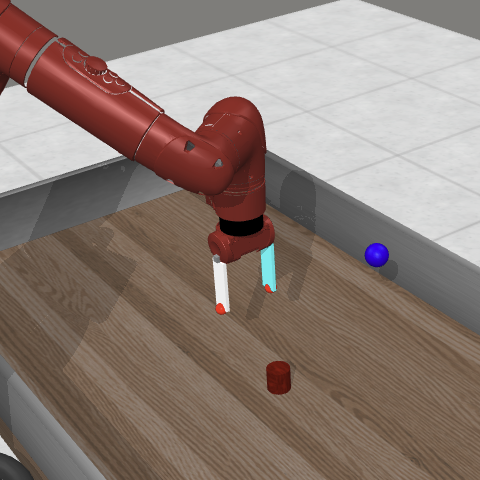}
\end{subfigure}  
\caption{Pick and Place}
\end{subfigure}
\hfill
\begin{subfigure}[b]{0.45\textwidth}
\begin{subfigure}[b]{0.45\textwidth}
    \centering
    \includegraphics[scale=0.2]{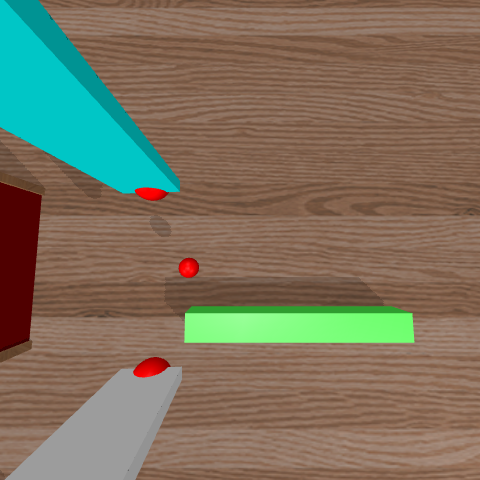}
\end{subfigure} 
\hfill
\begin{subfigure}[b]{0.45\textwidth}
    \centering
    \includegraphics[scale=0.2]{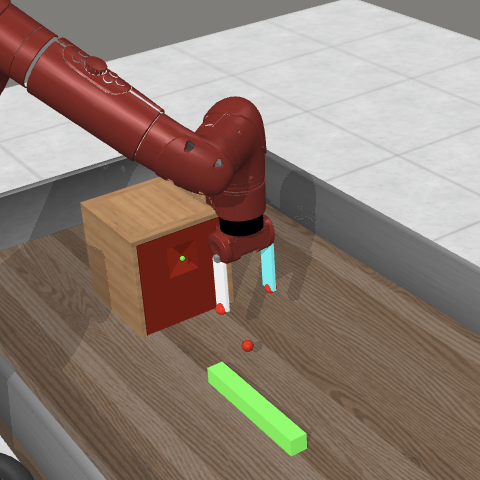}
\end{subfigure} 
\caption{Peg Insert}
\end{subfigure}
\caption{Images showing each camera view used for the MetaWorld tasks.}
\label{fig:all_metaworld_cams}
\end{figure}

We evaluate performance on four tasks from the MetaWorld benchmark suite \citep{Yu2020metaworld}. All tasks use the simulated Sawyer robotic arm and are implemented in the MuJoCo physics engine \citep{Todorov2012mujoco}. We use the `Multi-Task 1' setup which randomises the goal position for every episode within a single task. The action space consists of the 3D change in end-effector position and 1D gripper control. The goal of each task is described in Table~\ref{table:metaworld}. For all tasks, we render camera images from first-person and third-person perspectives for the observations, depicted in Figure~\ref{fig:all_metaworld_cams}.

\begin{table}[!h]
	\centering
	\begin{tabular}{ | c | c | }
		\hline
		Task & Goal \\ 
		\hline 
		Soccer & Guide the ball in to the goal \\
		Basketball & Pick up the basketball and place into the hoop \\
		Pick and Place & Move the object to the goal position \\
		Peg Insert & Pick up and insert a peg sideways into the hole \\ 
		\hline 
	\end{tabular}
	\vspace{1ex}
	\caption{Description of MetaWorld tasks.}
\label{table:metaworld}
\end{table}

\end{document}